\documentclass[lettersize,journal]{IEEEtran}
\usepackage{amsmath,amsfonts}
\usepackage{algorithmic}
\usepackage[ruled,linesnumbered]{algorithm2e}
\usepackage{array}
\usepackage[caption=false,font=normalsize,labelfont=sf,textfont=sf]{subfig}
\usepackage{textcomp}
\usepackage{stfloats}
\usepackage{url}
\usepackage{verbatim}
\usepackage{graphicx}
\usepackage{cite}
\usepackage{xspace}
\usepackage[capitalise]{cleveref}
\usepackage{wrapfig}
\usepackage{booktabs}
\usepackage[table]{xcolor}
\usepackage[inline]{enumitem}
\usepackage{multirow}
\usepackage{makecell}
\usepackage[caption=false,font=normalsize,labelfont=sf,textfont=sf]{subfig}
\usepackage{caption}
\usepackage{nameref}
\usepackage{pifont}%
\newcommand{\cmark}{\ding{51}}%
\newcommand{\xmark}{\ding{55}}%

\Crefname{section}{Sec.}{Secs.}
\Crefname{table}{Tab.}{Tabs.}
\Crefname{figure}{Fig.}{Figs.}
\Crefname{equation}{Eq.}{Eqs.}
\Crefname{algorithm}{Alg.}{Algs.}

\makeatletter
\DeclareRobustCommand\onedot{\futurelet\@let@token\@onedot}
\def\@onedot{\ifx\@let@token.\else.\null\fi\xspace}

\def\eg{\emph{e.g}\onedot} 
\def\ie{\emph{i.e}\onedot} 
 
\def\etc{\emph{etc}\onedot}

\makeatother

\begin{document}

\title{CAIT: Triple-Win \underline{C}ompression Towards High \underline{A}ccuracy, Fast \underline{I}nference, and Favorable \underline{T}ransferability for ViTs}

\author{Ao Wang, Hui Chen, Zijia Lin, Sicheng Zhao,~\IEEEmembership{Senior Member,~IEEE} \\Jungong Han,~\IEEEmembership{Senior Member,~IEEE}, Guiguang Ding,~\IEEEmembership{Senior Member,~IEEE}

\IEEEcompsocitemizethanks{\IEEEcompsocthanksitem A. Wang, H. Chen, S-C. Zhao and G-G. Ding are with BNRist, Tsinghua University, Beijing, China. A. Wang and G-G. Ding are also with School of Software, Tsinghua University, Beijing, China. (e-mail: wanga24@mails.tsinghua.edu.cn, jichenhui2012@gmail.com, schzhao@tsinghua.edu.cn, dinggg@tsinghua.edu.cn).\protect
\IEEEcompsocthanksitem Z-J. Lin is with School of Software, Tsinghua University, Beijing, China (e-mail: linzijia07@tsinghua.org.cn).
\IEEEcompsocthanksitem J-G. Han is with Department of Automation, Tsinghua University, Beijing, China. (e-mail: jungonghan77@gmail.com).
\protect
\IEEEcompsocthanksitem Corresponding authors: Hui Chen and Guiguang Ding.
}}

\markboth{IEEE Transactions on Pattern Analysis and Machine Intelligence,~Vol.~14, No.~8, August~2021}%
{Wang \MakeLowercase{\textit{et al.}}: CAIT: Triple-Win \underline{C}ompression Towards High \underline{A}ccuracy, Fast \underline{I}nference, and Favorable \underline{T}ransferability for ViTs}

\maketitle

\begin{abstract}
    Vision Transformers (ViTs) have emerged as state-of-the-art models for various vision tasks recently. However, their heavy computation costs remain daunting for resource-limited devices. To address this, researchers have dedicated themselves to compressing redundant information in ViTs for acceleration. However, existing approaches generally sparsely drop redundant image tokens by token pruning or brutally remove channels by channel pruning, leading to a sub-optimal balance between model performance and inference speed. Moreover, they struggle when transferring compressed models to downstream vision tasks that require the spatial structure of images, such as semantic segmentation. To tackle these issues, we propose CAIT, a joint \underline{c}ompression method for ViTs that achieves a harmonious blend of high \underline{a}ccuracy, fast \underline{i}nference speed, and favorable \underline{t}ransferability to downstream tasks. Specifically, we introduce an asymmetric token merging (ATME) strategy to effectively integrate neighboring tokens. It can successfully compress redundant token information while preserving the spatial structure of images. On top of it, we further design a consistent dynamic channel pruning (CDCP) strategy to dynamically prune unimportant channels in ViTs. Thanks to CDCP, insignificant channels in multi-head self-attention modules of ViTs can be pruned uniformly, significantly enhancing the model compression.  Extensive experiments on multiple benchmark datasets show that our proposed method can achieve state-of-the-art performance across various ViTs.
\end{abstract}

\begin{IEEEkeywords}
Model Compression, Vision Transformer, Channel Pruning, Token Pruning
\end{IEEEkeywords}

\section{Introduction}
\IEEEPARstart{R}{ecently}, the field of computer vision has witnessed significant progress with the emergence of Vision Transformer (ViT)~\cite{dosovitskiy2020image} and its variants~\cite{9716741,yao2023dual,yu2023metaformer,liu2021swin,touvron2021training}. These models have demonstrated exceptional performance on various vision tasks~\cite{xu2023vitpose,guo2022transformer,wu2022p2t,song2023vision,xiao2022image,chen2021pre}, surpassing the state-of-the-art convolutional neural networks (CNNs). Building upon the success of transformers~\cite{wei2022emergent,chowdhery2022palm,OpenAI_2023} in natural language processing (NLP), scaling ViTs has become a key priority~\cite{chen2022pali,riquelme2021scaling,zhai2022scaling,dehghani2023scaling}. This has led to the development of various vision foundation models, such as ViT-22B~\cite{dehghani2023scaling} and SAM~\cite{kirillov2023segment}. However, the high computation and memory costs of these models have posed significant challenges~\cite{chen2021chasing,kong2022spvit}, limiting their practical applications, especially on resource-limited devices. Therefore, compressing and accelerating ViTs are critical for making them viable for real-world applications~\cite{yu2022unified,zheng2022savit,rao2021dynamicvit}.

\setlength{\floatsep}{9pt}
\setlength{\textfloatsep}{5pt}

\begin{figure}
  \centering
  \includegraphics[width=0.85\linewidth]{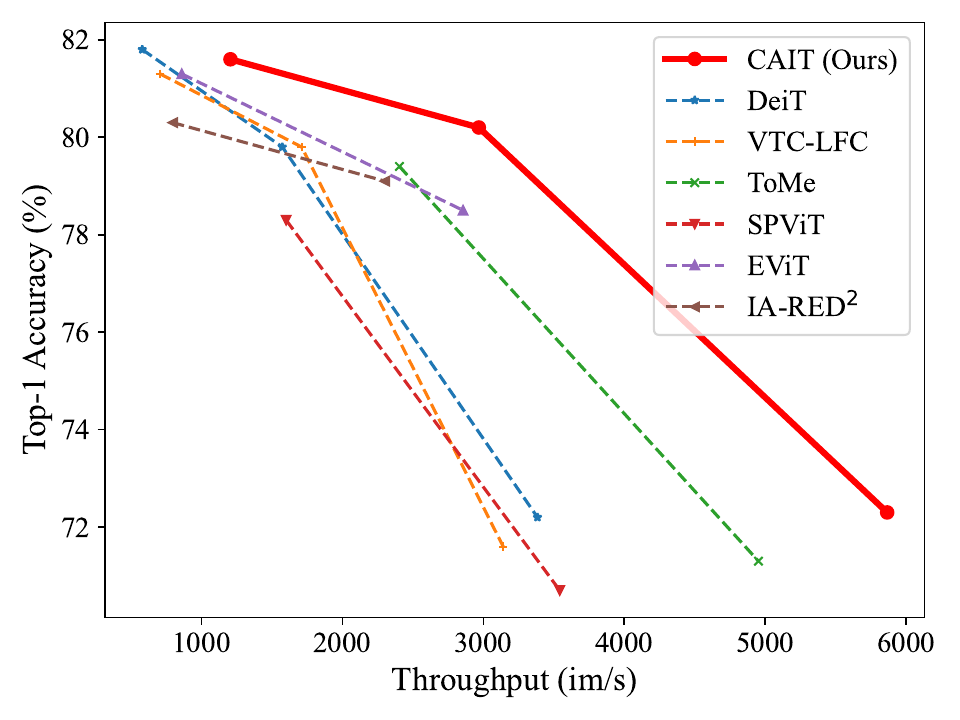}
  \caption{Comparison of throughput and accuracy between CAIT (Ours) and other models. The top-1 accuracy is tested on ImageNet-1K and the throughput is evaluated on a NVIDIA RTX-3090 GPU with a batch size of 256.}
  \label{fig:speed}
\end{figure}

Early attempts follow previous experiences in compressing CNN models, which aim to reduce redundant channels in a structured manner. They usually adopt a pruning-then-finetuning scheme via sparse learning~\cite{zhu2021vision}, taylor expansion~\cite{yang2021nvit,wang2022vtc}, or collaborative optimization~\cite{zheng2022savit}. Dynamic channel pruning~\cite{chen2021chasing,yu2022unified} is also applied for ViTs to identify unimportant channels during fine-tuning, achieving remarkable performance. Based on the intuition that many tokens encode less important or similar information, such as background details~\cite{tang2022patch,rao2021dynamicvit,kong2022spvit,liang2022not}, recent works investigate to prune redundant tokens to accelerate the transformer computation. For example, DynamicViT~\cite{rao2021dynamicvit} eliminates tokens based on their predicted importance scores. Considering that channel pruning and token pruning compress redundant information from model level (\ie, parameters) and data level (\ie, tokens), respectively, conducting them separately may lead to an excessive reduction on one level while neglecting the redundancy on the other level, which compromises overall model quality~\cite{chen2021chasing,wang2022vtc}. Thus, recently, there have been works utilizing token pruning and channel pruning for collaborative compression of ViTs, achieving state-of-the-art performance~\cite{hou2022multi,wang2022vtc}.

\begin{figure*}[t]
  \centering
  \includegraphics[width=0.85\textwidth]{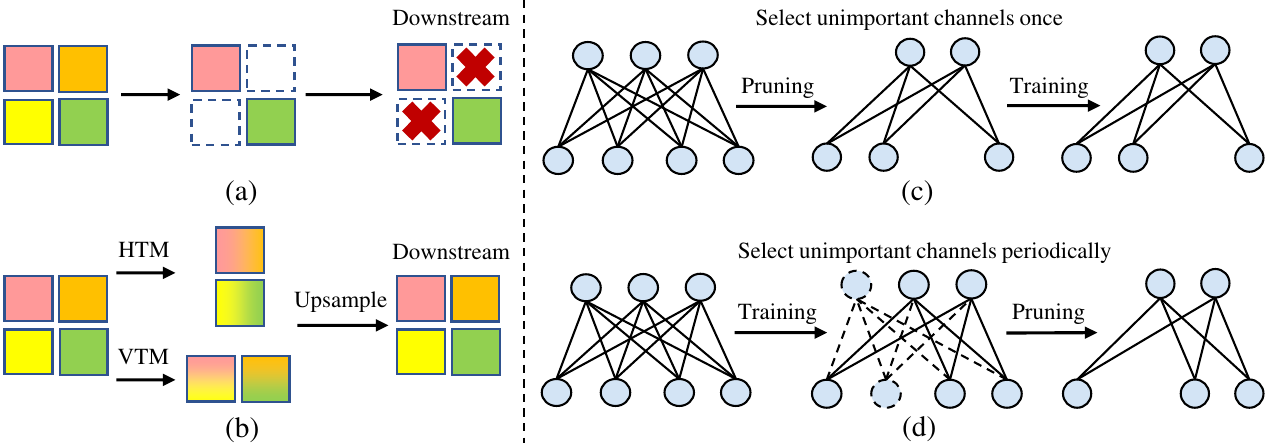}
  \caption{The comparisons between CAIT and others. (a) indicates that previous token pruning methods usually sparsely drop the tokens and disrupt the complete spatial structure when transferring to downstream tasks. (b) shows that ATME leverages horizontal token merging (HTM) and vertical token merging (VTM) to prune tokens while maintaining the spatial integrity, where image feature maps can be easily upsampled for downstream tasks. (c) denotes that previous channel pruning methods usually suffer from irreversible removal of important channels due to the direct pruning. (d) presents that instead of directly modifying the model structure like previous works, CDCP dynamically determines the importance of channels periodically and encourages structured sparsity network gradually during training, which can thus recover the important channels.}
  \label{fig:compare}
  \vspace{-12pt}
\end{figure*}

It is crucial to ensure that a compressed model can perform effectively and efficiently in practical applications. Therefore, existing works often prioritize compressing a pretrained model based on the principles of high accuracy and fast inference. High accuracy ensures that the capacity of the compressed model remains comparable to the original model. Fast inference, on the other hand, guarantees that the compressed model can make predictions quickly, which is particularly important in resource-constrained environments where low latency is essential. However, we argue that relying solely on accuracy and inference efficiency does not guarantee favorable transferability, \ie, enjoying the same remarkable performance in various downstream tasks. Existing works predominantly focus on achieving high accuracy in image classification on the ImageNet~\cite{deng2009imagenet} dataset without sufficient consideration for performance after transferring to downstream tasks. Therefore, they generally struggle to achieve triple-wins among the accuracy, the inference efficiency, and the transferability.
For example, advanced channel pruning methods~\cite{chen2021chasing,yu2022unified} directly remove attention heads without deeper exploration of sparsity in the multi-head self-attention (MHSA) module, easily causing over-pruning of parameters in MHSA. Unstructured token pruning methods \cite{rao2021dynamicvit,kong2022spvit,liang2022not,wang2022vtc,pan2021ia,bolya2022token,wei2023joint} usually drop redundant tokens sparsely, resulting in the disruption of the spatial structure of images, and thus leading to harmful impacts in downstream tasks heavily reliant on visual structure, such as semantic segmentation~\cite{xiao2018unified,cheng2022masked}, instance segmentation~\cite{he2017mask,cai2019cascade}, visual enhancement~\cite{cheng2019encoder}, and so on. Structured token pruning~\cite{xu2022evo,chang2023making} methods can maintain the spatial structure of images. However, they obtain inferior performance to unstructured ones~\cite{wang2022vtc,wei2023joint}. State-of-the-art methods~\cite{hou2022multi,wang2022vtc}, which combine token pruning and channel pruning, simply adopt principles of unstructured token pruning and pruning-then-finetuning channel pruning. They still fail to achieve a good balance among the performance, inference speed, and the transferability. For example, VTC-LFC~\cite{wang2022vtc} enjoys state-of-the-art performance but with slow inference speed and limited transferability.

In this work, we aim to deliver a triple-win \textbf{\underline{C}}ompression method, dubbed \textbf{CAIT}, which achieves \textbf{high \underline{A}ccuracy}, \textbf{fast \underline{I}nference speed}, and \textbf{favorable \underline{T}ransferability} all at once for pretrained ViTs. CAIT comprises two key strategies: the asymmetric token merging (ATME) strategy and the consistent dynamic channel pruning (CDCP) strategy. As shown in \Cref{fig:compare}.(a) and \Cref{fig:compare}.(b), rather than sparsely leaving out redundant tokens, ATME utilizes horizontal token merging and vertical token merging to integrate neighboring tokens. When adapting the model for downstream vision tasks after token pruning, the feature map of patches can be simply upsampled to restore the original spatial integrity. Therefore, ATME can effectively reduce the number of tokens while maintaining a complete spatial structure. Meanwhile, as shown in \Cref{fig:compare}.(c) and \Cref{fig:compare}.(d), CDCP adopts the dynamic channel pruning approach and encourages the structured sparsity model gradually by selecting unimportant channels periodically, avoiding the irreversible removal of important channels in previous works. It also employs head-level consistency and attention-level consistency to perform fine-grained compression for all modules. As a result, unimportant channels in ViTs, including the MHSA modules, can be uniformly removed, enabling fast parallel computing and thus enhancing the model compression. Besides, CDCP has no impact on the spatial structure of the image patches and preserves their spatial integrity. Combined with its high performance after pruning, it also facilitates the transferability of models. \Cref{tab:comp} presents the comparison of our proposed CAIT with other methods.

\begin{table}[t]
  \small
  \centering
  \caption{Comparison of the proposed CAIT with several state-of-the-art methods.}
  \label{tab:comp}
  \setlength\tabcolsep{8pt}%
  \resizebox{\linewidth}{!}{
  \begin{tabular}[t]{l|cccc}
    \toprule
    Method & \makecell{High \\accuracy} &  \makecell{Fast \\inference} & \makecell{Favorable \\transferability} \\
    \midrule
    EViT~\cite{liang2022not} & \cmark & \cmark & \xmark \\
    ToMe~\cite{bolya2022token} & \cmark & \cmark & \xmark \\
    Evo-ViT~\cite{xu2022evo} & \xmark & \cmark & \cmark \\
    IA-RED$^2$~\cite{pan2021ia} & \cmark & \cmark & \xmark \\
    dTPS~\cite{wei2023joint} & \cmark & \cmark & \xmark \\
    SPViT~\cite{he2021pruning} & \cmark & \xmark & \cmark \\
    VTC-LFC~\cite{wang2022vtc} & \cmark & \xmark & \xmark \\
    \rowcolor{lightgray}
    \textbf{CAIT (ours)} & \cmark & \cmark & \cmark \\
    \bottomrule
  \end{tabular}
  }
\end{table}

The proposed joint compression method can be seamlessly applied to prune well pretrained ViTs through a single fine-tuning process. Thanks to ATME and CDCP, redundant tokens and channels in pretrained ViTs can be simultaneously compressed, resulting in a considerable boost of computation efficiency without performance degradation. More importantly, the spatial structure of images are largely preserved during pruning, offering significant benefits for transferring to downstream tasks. Experiments on ImageNet show that our method can significantly outperform the state-of-the-art methods in terms of both the performance and the inference speed, as shown in \Cref{fig:speed}. Notably, our pruned DeiT-Tiny and DeiT-Small can achieve speedups of 1.7$\times$ and 1.9$\times$, respectively, without any compromise in performance. Our compressed DeiT-Base model achieves an impressive speedup of 2.1$\times$ with a negligible 0.2\% accuracy decline. In addition, when adapting our accelerated backbones to the downstream vision task of semantic segmentation, our method can provide up to 1.31$\times$ faster overall throughput without sacrificing performance, demonstrating its strong transferability.

In summary, our contributions are four-fold:
\begin{itemize}
  \item Beyond high accuracy and fast inference speed, we propose the incorporation of a novel principle: favorable transferability when designing compression algorithms. Thus, we present a joint compression method, dubbed CAIT, towards high accuracy, fast inference speed, and favorable transferability all at once for ViTs.
  \item We propose an asymmetric token merging strategy that effectively reduces the number of tokens while preserving complete spatial structure of images. It results in efficient models that are highly suitable for downstream tasks.
  \item We introduce consistent dynamic channel pruning strategy that achieves dynamic fine-grained compression optimization for all modules, further enhancing compression.
  \item Extensive experiments on various ViTs show that our method consistently achieves state-of-the-art results in terms of accuracy and inference speed, demonstrating its effectiveness. Experiments on transferring pruned ViTs to various downstream tasks verify the excellent transferability of our proposed method.
\end{itemize}

The subsequent sections of the paper are structured as follows. In \Cref{sec:related}, we conduct a thorough examination of related literature. Subsequently, in \Cref{sec:method}, we present our joint compression method CAIT, including the details of ATME and CDCP. In \Cref{sec:experiments}, we evaluate the performance of CAIT and conduct comprehensive analyses on various benchmark datasets. Finally, we conclude in \Cref{sec:conclusion}.

\section{Related work}
\label{sec:related}
\textbf{Vison Transformer.}
Inspired by remarkable achievements of transformer models~\cite{vaswani2017attention} in natural language processing, Vision Transformer (ViT)~\cite{dosovitskiy2020image} was introduced to leverage the pure transformer architecture for vision tasks.
With large-scale training data, ViT has shown outstanding performance on various image classification benchmarks, surpassing state-of-the-art convolutional neural networks (CNNs)~\cite{dosovitskiy2020image,touvron2021training,guo2022cmt}. Since then, many follow-up variants of ViT have been proposed~\cite{bao2021beit,yuan2021tokens,chen2021crossvit,ali2021xcit,graham2021levit,wang2022kvt,chen2021vision}. For example, DeiT~\cite{touvron2021training} presents a data efficiency training strategy for ViT by leveraging the teacher-student architecture.
In addition to image classification, many novel ViTs have also achieved remarkable performance in various other vision tasks, such as object detection~\cite{amini2022t6d,zhu2020deformable,dai2021up,misra2021end}, image retrieval~\cite{he2021transreid,el2021training}, semantic segmentation~\cite{cheng2021per,wang2021end,ding2106looking,zheng2021rethinking}, image reconstruction~\cite{chen2021pre,yang2020learning}, and 3D point cloud processing~\cite{lai2022stratified}. However, despite impressive performance, the intensive computation costs and memory footprint greatly hinder the efficient deployment of ViTs for practical applications~\cite{chen2021chasing}. This naturally calls for the study of efficient ViTs, including token pruning~\cite{ryoo2021tokenlearner,rao2021dynamicvit}, channel pruning~\cite{zhu2021vision,wang2022vtc},
and weights sharing~\cite{zhang2022minivit}, \etc.

\textbf{Token Pruning for ViTs.}
Token pruning for ViTs aims to reduce the number of processed tokens to accelerate the inference speed~\cite{rao2021dynamicvit,liang2022not}.
For example, DynamicViT~\cite{rao2021dynamicvit} removes less important tokens by evaluating their significance via a MLP based prediction module.
Additionally, SiT~\cite{zong2022self} proposes a token slimming module by dynamic token aggregation, meanwhile leveraging a feature distillation framework to recalibrate the unstructured tokens. 
Although achieving promising performance, most existing token pruning methods select tokens in an unstructured manner~\cite{rao2021dynamicvit,liang2022not,wang2022vtc,kong2022spvit,pan2021ia,bolya2022token,wei2023joint}, \ie, discarding redundant tokens sparsely, which inevitably damages the integrity of spatial structure. This greatly hinders the accelerated model transferred to downstream vision tasks depending on a complete spatial structure, such as semantic segmentation. Besides, existing structured token pruning methods~\cite{chang2023making,xu2022evo} fail to maintain dense pixel information or discriminative token features, still leading to the limited transferability.

\textbf{Channel Pruning for ViTs.}
Channel pruning for ViTs involves removing redundant parameters to obtain a more lightweight model~\cite{chen2021chasing}. For example,
NViT~\cite{yang2021nvit} proposes to greedily remove redundant channels by estimating their importance scores with the Taylor-based scheme. Additionally, SAViT~\cite{zheng2022savit} explores collaborative pruning by integrating essential structural-aware interactions between different components in ViTs. However, most channel pruning methods typically follow a two-stage approach and thus suffer from limitations stemming from the pruning process, in which the irreversible pruning may result in irreparable loss of important channels being erroneously pruned~\cite{zheng2022savit,wang2022vtc,yang2021nvit}. Besides, existing dynamic channel pruning methods for ViTs~\cite{chen2021chasing,yu2022unified} have limitations when it comes to the fine-grained pruning of MHSA modules due to the self-attention dimension constraints, thus leading to sub-optimal model quality and performance after compression.

\textbf{Joint Compression for ViTs.} Joint compression for ViTs aims to utilize token pruning and channel pruning for collaborative compression. It reduces both redundant data-level (\ie, tokens) and model-level (\ie, parameters) information in ViTs, achieving state-of-the-art performance~\cite{hou2022multi,wang2022vtc}. For example, VTC-LFC~\cite{wang2022vtc} presents bottom-up cascade pruning framework to jointly compress channels and tokens that are less effective to encode low-frequency information. ~\cite{hou2022multi} proposes a statistical dependence based pruning criterion to identify deleterious tokens and channels jointly. However, existing joint compression methods simply adopt the unstructured token pruning and pruning-then-finetuning channel pruning principles, failing to achieve superiority over model performance, the inference speed and transferability at the same time.

\section{Methodology}
\label{sec:method}

\begin{figure*}[t]
  \centering
  \includegraphics[width=0.85\textwidth]{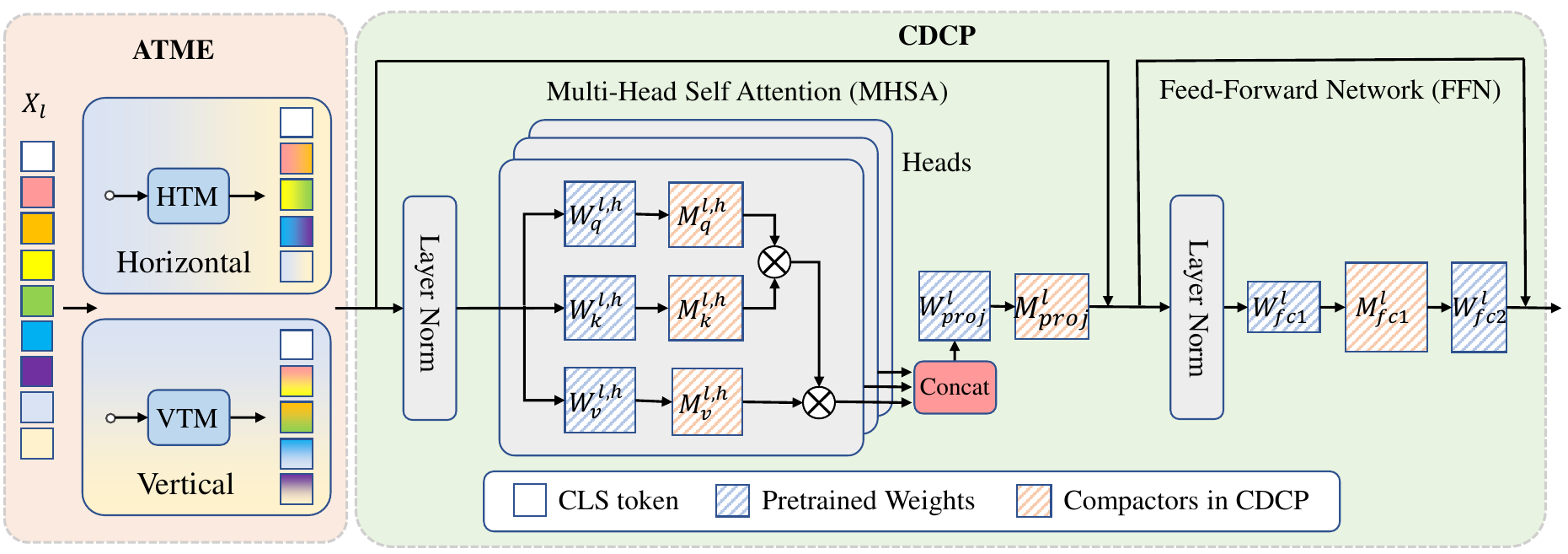}
  \caption{The overview of our proposed joint compression method for ViTs. We design an asymmetric token merging (ATME) strategy with horizontal token merging (HTM) and vertical token merging (VTM) to prune tokens while preserving their spatial integrity. Consistent dynamic channel pruning (CDCP) is further introduced to enable dynamic fine-grained compression optimization for all learnable weights with minimal performance degradation.}
  \label{fig:pipeline}
\end{figure*}

\begin{figure*}[!htb]
  \centering
  \includegraphics[width=0.85\linewidth]{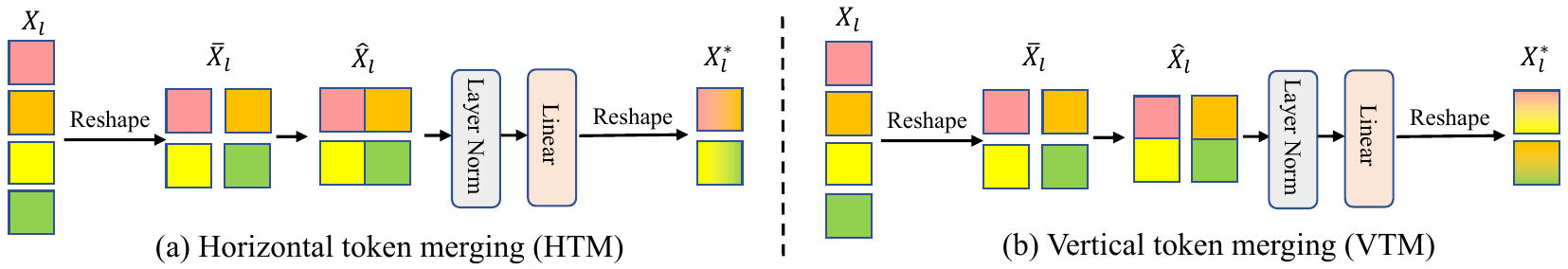}
  \caption{The proposed asymmetric token merging strategy prunes tokens horizontally or vertically.}
  \label{fig:horizontal}
  \vspace{-10pt}
\end{figure*}

\subsection{Preliminary}
We first introduce the necessary notations. The ViT model is composed of $L$ stacked transformer blocks. As shown in \Cref{fig:pipeline}, each transformer block comprises a multi-head self-attention (MHSA) module and a feed-forward network (FFN) module. For ease of explanation, we omit the CLS token for all input notations, because the CLS token is not involved in the token pruning. Given an input image, it is split into a sequence of tokens by \textsf{patchify} operation, which is then fed into transformer blocks to extract visual features. We use $X_l \in R^{N_l\times C}$ to denote the tokens in the $l$-th block, where $N_l$ is the number of tokens and $C$ is the dimension of token's feature. In the $l$-th transformer block, MHSA is parameterized by $W_{q}^{l,h}$, $W_{k}^{l,h}$, $W_{v}^{l,h} \in R^{C \times D}$ and $W_{proj}^l \in R^{C \times C}$, where $h$ denotes the index of head and $D$ is the head dimension. It can be formulated by:
\begin{equation}
\begin{split}
  \text{MHSA}(X_l) = \text{CONCAT}(\text{head}_0, ..., \text{head}_h, ...)W_{proj}^l, 
  \\
  \text{head}_h = \text{softmax}(\frac{(X_lW_q^{l,h})(X_lW_k^{l,h})^T}{\sqrt{D}})(X_lW_v^{l,h}).
\end{split}
\end{equation}
Similarly, FFN is parameterized by $W_{fc1}^l \in R^{C \times 4C}$ and $W_{fc2}^l \in R^{4C \times C}$. In this work, we aim to simultaneously reduce the token number $N_l$ and prune redundant channels in all parameters through the proposed joint compression method, as illustrated by \Cref{fig:pipeline}.

\subsection{Asymmetric Token Merging}
Most existing token pruning methods focus solely on image classification, and generally reduce the number of tokens in an unstructured manner~\cite{rao2021dynamicvit,kong2022spvit,liang2022not,wang2022vtc,pan2021ia,bolya2022token,wei2023joint}, \ie, by discarding tokens sparsely. However, although remarkable success has been achieved, sparsely wiping out redundant tokens inevitably disrupts the spatial integrity of images. Thus, the compressed ViT models are not suitable for downstream vision tasks that depend on a complete spatial structure, such as semantic segmentation, which significantly restricts their transferability. Besides, existing structured token pruning methods~\cite{chang2023making,xu2022evo} suffer from significant loss of dense information or token features, resulting in inferior performance during transferring. Here, we present an asymmetric token merging strategy to effectively accelerate ViTs, meanwhile maintaining the strong transferability of ViTs. Specifically, we introduce two basic token merging operations to integrate token features while preserving spatial integrity.

\textbf{Horizontal Token Merging (HTM)}. As shown in \Cref{fig:horizontal}.(a), for a sequence of tokens $X_l \in R^{N_l \times C}$ to be processed, we first reshape it to the shape of feature maps, \ie, $\overline{X}_l \in R^{H \times W \times C}$, where $H$ and $W$ are the height and width of features maps. Then, we group and concatenate two adjacent tokens horizontally, by which we can obtain $\widehat{X}_l \in R^{H \times \frac{W}{2} \times 2C}$, where $2C$ is the feature dimension after concatenation. We leverage a lightweight linear layer to effectively fuse the features of grouped tokens by $\widetilde{X}_l = \textsf{Linear}(\textsf{LayerNorm}(\widehat{X}_l)) \in R^{H \times \frac{W}{2} \times C}$. We then reshape it back to obtain a sequence of tokens, \ie, $X^*_l \in R^{\frac{N_l}{2} \times C}$.

\textbf{Vertical Token Merging (VTM)}. As shown in \Cref{fig:horizontal}.(b), similar to horizontal token merging, after obtaining $\overline{X}_l \in R^{H \times W \times C}$, we group and concatenate two adjacent tokens along the vertical direction, ending up with $\widehat{X}_l \in R^{\frac{H}{2} \times W \times 2C}$. Similarly, we can obtain the fused token features by $\widetilde{X}_l = \textsf{Linear}(\textsf{LayerNorm}(\widehat{X}_l)) \in R^{\frac{H}{2} \times W \times C}$. Then, we can derive the final token features $X^*_l \in R^{\frac{N_l}{2} \times C}$ after decreasing the number of tokens by reshaping.

By leveraging these two basic operations, we can obtain asymmetric feature maps through asymmetric merging in ViTs. Besides, both operations are generic and plug-and-play. We can seamlessly integrate them into ViTs without complicated hyper-parameter tuning. Following~\cite{rao2021dynamicvit,liang2022not,wei2023joint,kong2022spvit}, we hierarchically alternatively utilize horizontal and vertical token merging before MHSA through the whole network for the token pruning. Specifically, we first prioritize the strategy of uniformly dividing layers for pruning based on the expected FLOPs reduction. For example, if the target FLOPs reduction ratio for token pruning for a 12-layer DeiT-Small is 43.3\%, we initially select the 5-th layer and 9-th layer which are evenly sampled to perform HTM and VTM, respectively, resulting in a FLOPs reduction of 41.1\%. Either HTM first or VTM first makes a negligible difference according to our results. Subsequently, minor adjustments are made to the positions of the pruning layers, to align better with the desired ratio of FLOPs reduction. For example, we then adjust the pruning layer from the 9-th to the 8-th layer, resulting in an exact FLOPs reduction of 43.3\%. In this way, we can progressively reduce the number of tokens in ViTs while still maintaining the integrity of spatial structure for image features.

\subsection{Consistent Dynamic Channel Pruning}
\label{sec:cdcp}

\textbf{Dynamic Channel Pruning.} Two-stage channel pruning methods involve pruning a pretrained model and subsequently fine-tuning the pruned model~\cite{he2018soft,he2019filter}. However, these methods have limitations due to the pruning process, in which irreversible pruning can lead to the unintended removal of crucial channels and cause irreparable loss. In contrast, dynamic channel pruning dynamically determines the importance of channels during fine-tuning and encourages unimportant channels to gradually approach zero importance~\cite{ding2021resrep,lin2018accelerating,hou2022chex}. After fine-tuning, channels converging to zero importance are eliminated, resulting in the compressed model. In such a way, important channels can be recovered during training, thus leading to improved overall performance. Previous works for dynamic channel pruning focus on the design of metrics for deciding the importance of channels. Among them, compactor-based method~\cite{ding2021resrep} achieves the state-of-the-art performance for CNN pruning. Here, we propose a consistent dynamic channel pruning strategy based on~\cite{ding2021resrep} to perform fine-grained compression optimization for all modules in ViTs.

\begin{figure}[t]
  \centering
  \includegraphics[width=\linewidth]{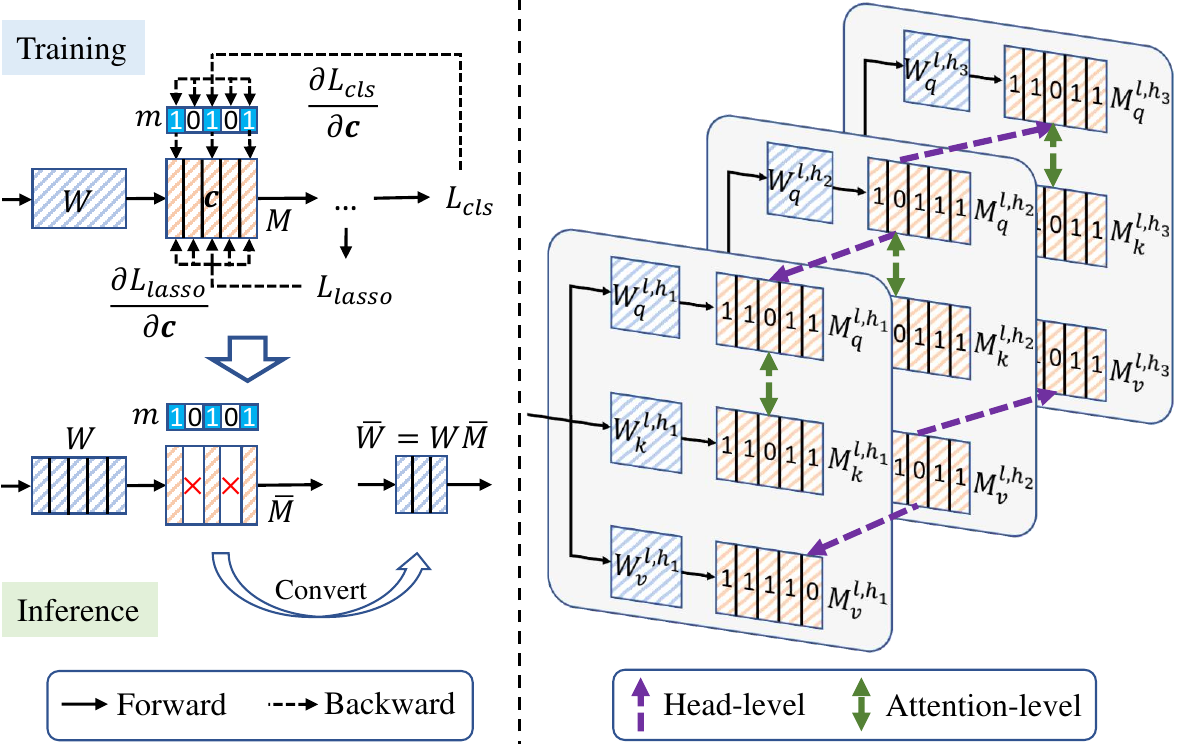}
  \caption{Left: Framework of the compactor. Right: Consistencies for pruning MHSA.}
  \label{fig:cdcp}
\end{figure}

As shown in \Cref{fig:cdcp}, following~\cite{ding2021resrep}, we insert a compactor, which is a learnable transformation matrix, for each parameter in ViT. For the generality of description, we denote a compactor and its preceding weight as $M$ and $W$, respectively, if not specified. Otherwise, we add super/sub-scripts to them to indicate their positions. For example, $M^{l,h}_q \in R^{D \times D}$ denotes the compactor corresponding to $W^{l,h}_q$ for the $h$-th head in the $l$-th block. Intuitively, in the compactor $M$, each column $\boldsymbol{c} \in M$ corresponds to one output channel of $W$. The norm of $\boldsymbol{c}$ can revel the importance of channels for $W$. Therefore, during training, we adopt the group lasso regularizer~\cite{liu2015sparse,wen2016learning} to dynamically push channels of $M$ to be sparse, \ie, $L_{lasso} = ||\boldsymbol{c}||_2$. As \cite{ding2021resrep}, we introduce a mask variable $m \in \{0, 1\}$ for each $\boldsymbol{c}$ to indicate the corresponding channel is pruned, \ie, $m=0$, or not,  \ie, $m=1$. We update the gradient of $\boldsymbol{c}$ manually by
\begin{equation}
  \bigtriangledown{c} = m\frac{\partial L_{cls}}{\partial \boldsymbol{c}} + \lambda \frac{\partial L_{lasso}}{\partial \boldsymbol{c}},
  \label{eq:cdcp_sparse}
\end{equation}
where $L_{cls}$ is the classification objective and $\lambda$ is a hyper-parameter. For every several iterations, we set the masks of $\boldsymbol{c}$ with the lowest norm values to 0 for encouraging the unimportant channels to approach zero importance. After training, we can eliminate redundant (converging to zero importance) channels in $M$, ending up with a pruned compactor $\overline{M}$. Then, a pruned weight, denoted as $\overline{W}$, can be derived by ${\overline{W}} = W {\overline{M}}$.

However, the vanilla compactor pruning applies a global selection criteria to identify unimportant channels and encourages them to approach zero importance. It may result in 
\begin{enumerate*}[label=(\arabic*)] 
  \item imbalanced sparse ratios of channels among heads after fine-tuning. Therefore, only the minimum ratio of near-zero importance channels across heads can be removed for efficient parallel self-attention computation; and 
  \item inconsistent channel importance between $W_q^{l,h}$ and $W_k^{l,h}$ which leads to different sparse outcomes. For example, one channel's importance may be close to zero importance in the query while its corresponding one in the key is not. Therefore, only the channels with zero importance in both the query and key can be pruned for error-free self-attention computation.
\end{enumerate*}
However, in such a way, a substantial number of near-zero importance channels are retained in the compressed model (\Cref{fig:consistency}), degrading the performance (\Cref{tab:cdcp}).

Here, to address these issues, we introduce the head-level consistency and attention-level consistency for pruning ViTs, as shown in \Cref{fig:cdcp}. Specifically, we first formulate the importance score of channel $\boldsymbol{c}$ as $s = ||\boldsymbol{c}||_2$. For channels with the same position in $M^{l,h}_q$ and $M^{l,h}_k$, their scores will be normalized as the mean of the corresponding original scores. Then, we can obtain a global set $S$, which contains scores of all channels in compactors. Meanwhile, we can derive local score sets, \ie, $S_q^{l,h}$, $S_k^{l,h}$ and $S_v^{l,h}$, for compactors in MHSA, \ie, $M_q^{l,h}$, $M_k^{l,h}$ and $M_v^{l,h}$, respectively. We initialize an empty set $P$ to record unimportant channels for sparsity via \Cref{eq:cdcp_sparse}. Then, we iteratively find unimportant channels and add them to $P$ until a pre-defined FLOPs reduction ratio $r_{target}$ is achieved. 

\textbf{Head-level Consistency}. We use this strategy to ensure that the ratios of unimportant channels across heads remain the same. It guarantees that after pruning, the same number of channels remain per head, allowing the MHSA module to be computed in parallel for fast inference. We take $M_q^{l, h}$ as an example. As shown in \Cref{alg:head}, suppose we have selected a channel $\boldsymbol{c} \in M_q^{l,h}$. For other head $h' \ne h$, we remove the smallest score from the local score set $S_q^{l, h'}$ (\Cref{alg:head3}) and add its corresponding channel $\boldsymbol{c}' \in M_q^{l,h'}$ to $P$ (\Cref{alg:head4}). At each iteration of channel selection of $\boldsymbol{c}$ in head $h$, we can make sure that the same channels can be added into $P$ for all heads, which will gradually become sparse during fine-tuning. After pruning, the same number of channels in different heads will thus be maintained, resulting in the same shape for each head.

\makeatletter
\let\original@algocf@latexcaption\algocf@latexcaption
\long\def\algocf@latexcaption#1[#2]{%
  \@ifundefined{NR@gettitle}{%
    \def\@currentlabelname{#2}%
  }{%
    \NR@gettitle{#2}%
  }%
  \original@algocf@latexcaption{#1}[{#2}]%
}
\makeatother

\begin{algorithm}[t]
    \caption{Head-level consistency}
    \label{alg:head}
    \KwIn{Selected channel $\boldsymbol{c}\in M_q^{l,h}$, score sets for each head $\{S_q^{l,1},..., S_q^{l, h}, ...\}$}
    \KwOut{Expanded channels $P$ for sparsity}
    \SetAlgoNoEnd
    $S_{q}^{l,h} \gets S_{q}^{l,h} \setminus \{s_{\boldsymbol{c}}\}$; $P \gets \{\boldsymbol{c}\}$\;
    \For{each head $h' \ne h$ } {
        $\boldsymbol{c}' \gets \mathop{\arg\min}_{\boldsymbol{c}' \in M_{q}^{l,h'}}(S_{q}^{l,h'})$\; \label{alg:head3}
        $S_{q}^{l,h'} \gets S_{q}^{l,h'} \setminus \{s_{\boldsymbol{c}'}\}$; $P \gets P \cup \{\boldsymbol{c}'\}$; \label{alg:head4}
    }
\end{algorithm}

\begin{algorithm}[t]
    \caption{Attention-level consistency}
    \label{alg:attention}
    \KwIn{Selected channel $\boldsymbol{c}\in M_q^{l,h}$ , score sets $\{S_q^{l, h}, S_k^{l, h}\}$}
    \KwOut{Expanded channels $P$ for sparsity}
    $S_{q}^{l,h} \gets S_{q}^{l,h} \setminus \{s_{\boldsymbol{c}}\}$; $P \gets \{\boldsymbol{c}\}$\;
    $i = index(M_{q}^{l,h}, \boldsymbol{c})$; 
    $\boldsymbol{c}' \gets M_{k}^{l,h}[i]$\;
    $S_{k}^{l,h} \gets S_{k}^{l,h} \setminus \{s_{\boldsymbol{c}'}\}$;
    $P \gets P \cup \{\boldsymbol{c}'\}$; 
\end{algorithm}

\textbf{Attention-level Consistency}. It is designed to encourage consistent behavior for channels in the same position between $M_q^{l,h}$ and $M_k^{l,h}$. As shown in \Cref{alg:attention}, if the $i$-th channel in $M_q^{l,h}$, \ie, $\boldsymbol{c}$, is added into $P$, the $i$-th channel in $M_k^{l,h}$, \ie, $\boldsymbol{c}'$, is added into $P$ as well, no matter how large its importance score is. Channels in $M_k^{l,h}$ will be managed in the same way. As a result, channels in query and key will be encouraged sparsity simultaneously. Therefore, after finetuning, channels with the same position in query and key will be of similar importance, thus being pruned or maintained consistently. This greatly benefit error-free interaction for the attention.

\textbf{Pipeline.} \Cref{alg:consistency} illustrates the proposed consistent dynamic channel pruning process. During finetuning, unconstrained channels, \ie, $\boldsymbol{c} \in M_{proj}^l \cup M_{fc1}^l$, are directly added to $P$. For channel $\boldsymbol{c} \in M_q^{l,h} \cup M_k^{l,h}$, we apply the proposed head-level consistency to it. After that, we employ the attention-level consistency to the newly to-be-sparse channels. For $\boldsymbol{c} \in M_v^{l,h}$, we only apply head-level consistency. After finetuning, channels of zero importance in $M$ are pruned, resulting in the pruned compactor $\overline{M}$. $\overline{M}$ can thus be merged with its preceding weight $W$ to obtain the compressed model.

\begin{algorithm}[t]               %
    \SetAlgoNoEnd
    \caption{Channel pruning with consistencies}
    \label{alg:consistency}
    \KwIn{Global importance score set $S$, target FLOPs reduction ratio $r_{target}$}
    \KwOut{Channels P for sparsity, current FLOPs reduction ratio $r_{current}$}
    \While{$r_{current} < r_{target}$} {
        $\boldsymbol{c} \gets \mathop{\arg\min}_{c}{S}$\;
        \uIf{$\boldsymbol{c} \in M_{proj}^l \cup M_{fc1}^l$}{
        $P \gets P \cup \{\boldsymbol{c}\}$; 
        $S \gets S \setminus \{s_{\boldsymbol{c}}\}$;
        }
        \uElseIf{$\boldsymbol{c} \in M_{v}^{l,h}$}{
        $P_{head} \gets$ \nameref{alg:head}($\boldsymbol{c}$)\;
        $P \gets P \cup P_{head}$; 
        $S \gets S \setminus S_{P_{head}}$;
        }
        \uElseIf{$\boldsymbol{c} \in M_{q}^{l,h} \cup M_{k}^{l,h}$} {
        $P_{head} \gets$ \nameref{alg:head}($\boldsymbol{c}$)\;
        \For{each channel $\boldsymbol{c}' \in P_{head}$}{
            $P_{attn} \gets$ \nameref{alg:attention}($\boldsymbol{c}'$)\;
            $P \gets P \cup P_{attn}$; 
            $S \gets S \setminus S_{P_{attn}};$
        }
        }
        Update $r_{current}$
    }
\end{algorithm}

\section{Experiments}
\label{sec:experiments}
We first compare our method with state-of-the-arts on ImageNet~\cite{deng2009imagenet} to verify the high performance and fast inference speed obtained by our method (\Cref{sec:imagenet}), following~\cite{wang2022vtc,zheng2022savit,yu2022unified}. We then evaluate the performance of our accelerated backbones on various downstream pixel-level vision tasks to demonstrate their strong transferability (\Cref{sec:transferability}). Additionally, we investigate impacts of each component by comprehensive analyses on ImageNet, following~\cite{wang2022vtc} (\Cref{sec:analyses}). Furthermore, we provide more evaluation, analyses, and visualization in \Cref{sec:discussion}. The float operations (FLOPs) of models are measured by fvcore~\cite{fvcore} and the throughput is evaluated on a single NVIDIA RTX-3090 GPU with a batch size of 256, by default. For compared methods, we utilize their published pretrained models to obtain throughputs.

\begin{table}[t]
  \small
  \centering
  \caption{Hyper-parameters during fine-tuning}
  \label{tab:parameter}
  \begin{tabular}{cc}
    \toprule
    \textbf{hyper-parameter} & \textbf{value} \\
    \midrule
    optimizer & AdamW \\
    base learning rate & 1e-4 \\
    weight decay & 0.05 \\
    optimizer momentum & \makecell{$\beta_1$ = 0.99 (compactor), 0.9 (other) \\ $\beta_2$ = 0.999} \\
    batch size & 256 (Tiny/Small), 128 (Base) \\
    learning rate schedule & cosine decay \\
    label smoothing & 0.1 \\
    mixup & 0.8 \\
    cutmix & 1.0 \\
    distillation loss & cross entropy loss \\
    distillation-alpha & 0.1 (Tiny), 0.25 (Small/Base) \\
    \bottomrule
  \end{tabular}
\end{table}

\newsavebox\dummy
\newcolumntype{H}{>{\begin{lrbox}{\dummy}}c<{\end{lrbox}}@{}}

\begin{table}[t]
    \caption{Comparison with state-of-the-arts on ImageNet.}
    \label{tab:imagenet}
    \small
    \setlength\tabcolsep{3pt}%
    \centering
    \resizebox{\linewidth}{!}{
    \begin{tabular}{lccHHcc}
      \toprule
      Model  & Param. (M) & FLOPs (G) & FLOPs ($\downarrow$\%) & Throughput (im/s) & Speed ($\uparrow$) & Top-1 (\%) \\
      \midrule
      DeiT-Tiny  & 5.7 & 1.3 & - & 3384 & 1.0$\times$ & 72.2 \\
      S$^2$ViTE~\cite{chen2021chasing}  & 4.2 & 1.0 & 23.7 & - & - & 70.1\\
      SPViT~\cite{he2021pruning}  & 4.9 & 1.0 & 23.1 & 3542 & 1.1$\times$ & 70.7\\
      ToMe~\cite{bolya2022token} & 5.7 & 0.7 & 41.3 & 4952 & 1.5$\times$ & 71.3 \\
      UVC~\cite{yu2022unified}  & - & 0.6 & 50.8 & - & - & 71.3 \\
      SAViT~\cite{zheng2022savit}  & 4.2 & 0.9 & 25.2 & - & - & 70.7 \\
      VTC-LFC~\cite{wang2022vtc} & 5.1 & 0.7 & 46.7 & 3141 & 0.9$\times$  & 71.6\\
      \rowcolor{lightgray}
      \textbf{CAIT (ours)}  & 5.1 & \textbf{0.6} & \textbf{50.5} & 5865 & \textbf{1.7$\times$} & \textbf{72.3} \\
      \midrule
      DeiT-Small  & 22.1 & 4.6 & - & 1574 & 1.0$\times$ & 79.8\\
      CP-ViT~\cite{song2022cp}  & 22.1 & 2.7 & 42.2 & - & - & 79.1\\
      EViT~\cite{liang2022not}  & 22.1 & 2.3 & 50.0 & 2856 & 1.8$\times$ & 78.5 \\
      IA-RED$^2$~\cite{pan2021ia}  & 22.1 & 3.2 & 31.5 & 2301 & 1.5$\times$ & 79.1\\
      dTPS~\cite{wei2023joint} & 22.8 & 3.0 & 34.8 & 2339 & 1.5$\times$ & 80.1 \\
      S$^2$ViTE~\cite{chen2021chasing}  & 14.6 & 3.2 & 31.6 & - & - & 79.2\\
      SPViT~\cite{he2021pruning}  & 16.4 & 3.3 & 28.3 & 1601 & 1.0$\times$ & 78.3\\
      ToMe~\cite{bolya2022token} & 22.1 & 2.7 & 41.3 & 2405 & 1.5$\times$ & 79.4 \\
      UVC~\cite{yu2022unified}  & - & 2.3 & 49.6 & - & - & 78.8\\
      SAViT~\cite{zheng2022savit} & 14.7 & 3.1 & 33.5 & - & -  &  80.1 \\
      VTC-LFC~\cite{wang2022vtc} & 17.7 & 2.1 & 54.4 & 1711 & 1.1$\times$ & 79.8\\
      \rowcolor{lightgray}
      \textbf{CAIT (ours)}  & 18.4 & \textbf{2.1} & \textbf{54.4} & 2968 & \textbf{1.9$\times$} & \textbf{80.2}\\
      \midrule
      DeiT-Base & 86.4 & 17.6 & - & 580 & 1.0$\times$ & 81.8 \\
      EViT~\cite{liang2022not}  & 86.4 & 11.6 & 34.1 & 860 & 1.5$\times$ & 81.3\\
      IA-RED$^2$~\cite{pan2021ia}  & 86.4 & 11.8 & 33.0 & 797 & 1.4$\times$ & 80.3\\
      S$^2$ViTE~\cite{chen2021chasing}  & 56.8 & 11.8 & 33.1 & - & - & 82.2 \\
      SPViT~\cite{he2021pruning}  & 62.3 & 11.7 & 33.1 & 600 & 1.0$\times$ & 81.6 \\
      UVC~\cite{yu2022unified}  & - & 8.0 & 54.4 & - &  - & 80.6 \\
      VTC-LFC~\cite{wang2022vtc}  & 63.5 & 7.5 & 57.6 & 707 & 1.2$\times$ & 81.3 \\
      \rowcolor{lightgray}
      \textbf{CAIT (ours)}  & 71.3 & \textbf{7.4} & \textbf{58.0} & 1206 & \textbf{2.1$\times$} & \textbf{81.6} \\
      \bottomrule
    \end{tabular}
    }
\end{table}

\subsection{Comparison Results on ImageNet}
\label{sec:imagenet}
\subsubsection{Implementation Details} We evaluate our proposed method on three different sizes of DeiT~\cite{touvron2021training}, \ie, DeiT-Tiny, DeiT-Small, and DeiT-Base. 
Our experiments are deployed with Pytorch~\cite{paszke2017automatic} on RTX-3090 GPUs. In CDCP, following~\cite{ding2021resrep}, $r_{target}$ is initialized to zero, which is then increased by 0.025\% every 25 iterations until achieving the given reduction ratio. Meanwhile, we re-construct $P$ every same iterations. Besides, we start to increase $r_{target}$ and re-construct $P$ after 30 epochs. $\lambda$ in \Cref{eq:cdcp_sparse} is empirically set to 1e-5. \Cref{tab:parameter} reports the detailed hyper-parameters during training, most of which are the same as \cite{wang2022vtc}. Following~\cite{wang2022vtc,zheng2022savit,yu2022unified}, the corresponding original models are used for hard distillation.

\subsubsection{Results} As shown in \Cref{tab:imagenet}, our proposed method can consistently outperform previous methods across all three models, as evidenced by superior performance in terms of the Top-1 accuracy, the FLOPs reduction ratio, and the inference speed. Specifically, under similar FLOPs reduction ratios, our method outperforms the state-of-the-art VTC-LFC~\cite{wang2022vtc} by 0.7\% and 0.4\% in terms of Top-1 accuracy on DeiT-Tiny and DeiT-Small, respectively, all while achieving significantly faster inference speeds. Such improvements can be attributed to the effective token merging and channel pruning by our ATME and CDCP, respectively. Compared with methods that obtain comparable accuracy to ours, such as dTPS~\cite{wei2023joint} and SPViT~\cite{he2021pruning}, our method can achieve much higher FLOPs reductions. We can see that in the proposed method, the fruitful FLOPs reduction can be sufficiently transformed into the significant inference acceleration. Notably, our compressed DeiT-Tiny, DeiT-Small, and DeiT-Base models can achieve 1.7$\times$, 1.9$\times$, and 2.1$\times$ inference speedups, respectively, while enjoying no or little accuracy drops. These results well demonstrate the effectiveness and the superiority of our method.

\subsection{Transferability on Downstream Pixel-Level Tasks}
\label{sec:transferability}

\begin{table*}[t]
  \centering
  \caption{Results on semantic segmentation.}
  \label{tab:segmentation}
  \resizebox{\linewidth}{!}{
  \begin{tabular}{c|ccc|ccc|ccc}
    \toprule
     & \multicolumn{3}{c|}{Semantic FPN~\cite{kirillov2019panoptic}} & \multicolumn{3}{c|}{UperNet~\cite{xiao2018unified}} & \multicolumn{3}{c}{Mask2Former~\cite{cheng2022masked}} \\
    \midrule
    Backbone & mIoU & \makecell{En. sp. $\uparrow$} & \makecell{Over. sp. $\uparrow$} & mIoU & \makecell{En. sp. $\uparrow$} & \makecell{Over. sp. $\uparrow$} & mIoU & \makecell{En. sp. $\uparrow$} & \makecell{Over. sp. $\uparrow$} \\
    \midrule
    DeiT-Small & 44.3 & 1.00$\times$ & 1.00$\times$ & 44.9 & 1.00$\times$ & 1.00$\times$ & 47.2 & 1.00$\times$ & 1.00$\times$\\
    \makecell{VTC-LFC-unstructured~\cite{wang2022vtc}} & 0.1 & 1.01$\times$ & 1.01$\times$ & 0.1 & 1.01$\times$ & 1.01$\times$ & 45.2 & 1.16$\times$ & 1.05$\times$ \\
    \makecell{VTC-LFC-structured~\cite{wang2022vtc}} & 43.5 & 1.01$\times$ & 1.00$\times$ & 44.2 & 1.00$\times$ & 1.00$\times$ & 46.3 & 1.15$\times$ & 1.05$\times$ \\
    Evo-ViT~\cite{xu2022evo} & 42.7 & 1.34$\times$ & 1.20$\times$ & 43.0 & 1.33$\times$  & 1.18$\times$  & 45.9 & 1.41$\times$ & 1.14$\times$\\
    \rowcolor{lightgray}
    \textbf{ATME} & \textbf{44.5} & \textbf{1.46$\times$} & \textbf{1.27$\times$} & \textbf{45.3} & \textbf{1.43$\times$} & \textbf{1.23$\times$} & \textbf{47.6} & \textbf{1.54$\times$} & \textbf{1.18$\times$} \\
    \rowcolor{lightgray}
    \textbf{CAIT} & \textbf{44.5} & \textbf{1.52$\times$} & \textbf{1.31$\times$}& \textbf{45.6} & \textbf{1.48$\times$} & \textbf{1.26$\times$} & \textbf{47.2} & \textbf{1.64$\times$} & \textbf{1.21$\times$} \\
    \bottomrule
  \end{tabular}
  }
  \vspace{-7pt}
\end{table*}

\subsubsection{Results on Semantic Segmentation}
Most existing token pruning methods generally reduce the number of tokens in an unstructured manner~\cite{rao2021dynamicvit,kong2022spvit,liang2022not,wang2022vtc,pan2021ia,bolya2022token,wei2023joint}, \ie, by dropping tokens sparsely, which however inevitably disrupts the complete spatial structure of images. Therefore, the accelerated ViTs are not suitable for downstream pixel-level vision tasks, like semantic segmentation. To verify the impact of unstructured token pruning on downstream vision tasks, we conduct experiments with the start-of-the-art VTC-LFC~\cite{wang2022vtc} on the ADE20k~\cite{zhou2017scene} dataset. Additionally, in contrast, our proposed method can preserve the spatial integrity and effectively adapt to downstream tasks that need a complete spatial structure of images. Therefore, we also conduct experiments on the ADE20k dataset to verify such a transferability. We introduce state-of-the-art Evo-ViT~\cite{xu2022evo} as one baseline, which can also maintain spatial structure of input images as ours.

Following~\cite{ali2021xcit,jiang2021all}, we integrate accelerated backbones into three advanced segmentation methods, \ie, Semantic FPN~\cite{kirillov2019panoptic}, UperNet~\cite{xiao2018unified},  and Mask2Former~\cite{cheng2022masked}. We train for 80k, 160k, and 160k iterations for these three segmentation methods, respectively. Besides, we adopt the AdamW~\cite{loshchilov2017decoupled} optimizer with the learning rate of 6e-5 and weight decay of 0.01, as in~\cite{liu2021swin}. The input resolution is set to 512$\times$512 and all models are trained using batch size of 32. We report the performance with standard single scale protocol as in ~\cite{ali2021xcit,jiang2021all}. Additionally, the encoder speedup (En. sp.) and overall speedup (Over. sp.) are evaluated on a single RTX-3090 GPU with a batch size of 32, where the encoder contains the backbone, upsampling and downsampling modules. Our implementation is based on \textsf{mmsegmentation}~\cite{contributors2020mmsegmentation}.

As VTC-LFC~\cite{wang2022vtc} produces sparse feature maps, following \cite{he2022masked,tian2023designing}, we use mask tokens to fill the dropped positions before feeding them into the semantic segmentation decoder, which is denoted as ``VTC-LFC-unstructured''. As shown in \Cref{tab:segmentation}, due to the impaired spatial integrity of feature maps, CNN-based decoders, \ie, Semantic FPN\cite{kirillov2019panoptic} and UperNet~\cite{xiao2018unified}, result in poor results. It is consistent with observations in previous works~\cite{tian2023designing,pathak2016context} that CNNs exhibit significantly worse performance when dealing with sparse feature maps, which can be attributed to the disrupted data distribution of pixel values and vanished patterns of visual representations. Besides, with the Transformer-based decoder, \ie, Mask2Former~\cite{cheng2022masked}, ``VTC-LFC-unstructured'' demonstrates a significant inferiority to DeiT-Small, with a considerable margin of 2.0\% mIoU. These results well show the harmful impacts caused by unstructured token pruning when transferring the accelerated model to downstream structured vision task of semantic segmentation. Furthermore, we propose to record the dropped tokens and then use them to fill the corresponding positions when constructing feature maps to be fed into the segmentation decoder, thus ensuring the spatial integrity of patches, which is denoted as ``VTC-LFC-structured''. As shown in \Cref{tab:segmentation}, reasonably, ``VTC-LFC-structured'' outperforms ``VTC-LFC-unstructured'' across three segmentation methods.

Furthermore, as shown in \Cref{tab:segmentation}, our method not only exhibits superior performance but also boasts fast inference 
speed across all semantic segmentation methods. Specifically, our ATME yields impressive overall speedups of 1.27$\times$, 1.23$\times$, and 1.18$\times$, respectively, across three distinct segmentation methods, while maintaining optimal performance. Our ATME outperforms ``VTC-LFC-structured'' by great margins of 1\%, 1.1\%, 1.3\% mIoUs on all segmentation decoders, respectively, with notably faster inference speedup. Besides, our ATME significantly outperforms Evo-ViT~\cite{xu2022evo} with 1.8\%, 2.3\%, and 1.7\% mIoU on three segmentation heads, respectively. It well indicates the superiority of asymmetric token merging in preserving spatial integrity, compared with Evo-ViT that can potentially harm token features. Besides, on top of ATME, our CAIT can further enhance the overall inference speed. These results well demonstrate the remarkable adaptability of the proposed method to downstream vision tasks.

\subsubsection{Results on Object Detection \& Instance Segment.}
To verify the strong transferability of our method on various downstream pixel-level vision tasks, we experiment over COCO-2017~\cite{lin2014microsoft} to evaluate the performance on object detection and instance segmentation. Following~\cite{wang2021pyramid}, we integrate the accelerated backbones into Mask-RCNN~\cite{he2017mask}. We adopt the AdamW optimizer with a initial learning rate of 2$\times$10$^{-4}$. The models are trained for 12 epochs with the input resolution of 1333$\times$800. We introduce the state-of-the-art ``VTC-LFC-structured'' as the baseline method.

As shown in \Cref{tab:instance}, thanks to preserving the complete spatial structure of image patches, our ATME significantly outperforms ``VTC-LFC-structured'' by 1.2 AP$^{box}$ and 0.8 AP$^{mask}$, respectively. Besides, our CAIT achieves 1.52$\times$ speedup without performance degradation, well demonstrating the strong transferability of our method.

\begin{table}[t]
  \small
  \centering
  \caption{Results on object detection \& instance segment.}
  \label{tab:instance}
  \begin{tabular}[t]{l|cccc}
    \toprule
    Method & AP$^{box}$ & \makecell{AP$^{mask}$} & \makecell{ Speed $\uparrow$ } \\
    \midrule
    DeiT-Small & 38.4 & 35.2 & 1.00$\times$ \\
    VTC-LFC-structured & 37.5 & 34.4 & 1.01$\times$ \\
    \rowcolor{lightgray}
    \textbf{ATME} & \textbf{38.7} & \textbf{35.2}  & \textbf{1.46$\times$} \\
    \rowcolor{lightgray}
    \textbf{CAIT} & \textbf{38.8} & \textbf{35.6} & \textbf{1.52$\times$} \\
    \bottomrule
  \end{tabular}
\end{table}

\subsubsection{Results on Medical and Aerial Segmentation}
\label{sec:medical}
In order to demonstrate the generalizability of our accelerated backbones to out-of-domain downstream tasks, we conduct experiments on medical image segmentation and aerial semantic segmentation tasks. Specifically, for medical image segmentation, following~\cite{rahman2023medical}, we integrate accelerated backbones into the CASCADE framework~\cite{rahman2023medical} and evaluate the performance on widely used Synapse multi-organ dataset~\cite{landman2015miccai}, ACDC dataset~\cite{bernard2018deep} and Polyp datasets~\cite{bernal2015wm,jha2020kvasir}. We follow~\cite{rahman2023medical} to report DICE score for all datasets. Regarding to aerial semantic segmentation, we follow~\cite{wang2022empirical} to adopt UperNet~\cite{xiao2018unified} as the unified segmentation framework. Besides, we conduct experiments on ISPRS Potsdam dataset~\cite{potsdam} and the large-scale segmentation benchmark, \ie, iSAID~\cite{waqas2019isaid}. We follow~\cite{wang2022empirical} to report the overall accuracy (OA) and mean F1 score (mF1) for the Potsdam dataset, and mIoU for the iSAID dataset. We introduce state-of-the-art ``VTC-LFC-structured'' as baseline.

As shown in \Cref{tab:medical}, benefiting from the preserved spatial integrity, our CAIT significantly outperforms ``VTC-LFC-structured'' by 2.3, 0.8 and 13.0 DICE score on Synapse, ACDC and Polyp datasets, respectively. Meanwhile, compared with DeiT-Small, it demonstrates 1.52$\times$ speedup with comparable performance across all three datasets. Similarly, as shown in \Cref{tab:aerial}, our CAIT significantly surpasses ``VTC-LFC-structured'' with 0.3 mF1 and 2.4 mIoU on Potsdam and iSAID, respectively. Besides, it attains a notable speed improvement of 1.52$\times$ without compromise in performance. Overall, thanks to well-preserved spatial integrity and dynamic fine-grained compression optimization by ATME and CDCP, respectively, our CAIT demonstrates robust generalizability across out-of-domain downstream tasks. 

\begin{table}[t]
  \small
  \centering
  \caption{Results on medical image segmentation.}
  \label{tab:medical}
  \resizebox{\linewidth}{!}{
  \begin{tabular}[t]{l|ccccc}
    \toprule
    Method & \makecell{Synapse} & \makecell{ACDC} & \makecell{Polyp} & \makecell{ Speed $\uparrow$ } \\
    \midrule
    DeiT-Small & 76.5 & 88.6 & 78.9  & 1.00$\times$ \\
    VTC-LFC-structured & 74.1 & 87.6 & 65.9 & 1.01$\times$ \\
    \rowcolor{lightgray}
    \textbf{CAIT} & \textbf{76.4} & \textbf{88.4} & \textbf{78.9} & \textbf{1.52$\times$} \\
    \bottomrule
  \end{tabular}
  }
\end{table}

\begin{table}[t]
  \small
  \centering
  \caption{Results on aerial semantic segmentation.}
  \label{tab:aerial}
  \setlength\tabcolsep{7pt}%
  \resizebox{\linewidth}{!}{
  \begin{tabular}[t]{l|cc|c|cc}
    \toprule
    \multirow{2}{*}{Method} & \multicolumn{2}{c|}{Potsdam} & iSAID & \multirow{2}{*}{Speed $\uparrow$} \\
    \cmidrule{2-4}
    & OA & mF1 & mIoU \\
    \midrule
    DeiT-Small & 88.6 & 90.9 & 60.8  & 1.00$\times$ \\
    VTC-LFC-structured & 88.2 & 90.6 & 58.7 & 1.01$\times$ \\
    \rowcolor{lightgray}
    \textbf{CAIT} & \textbf{88.5} & \textbf{90.9} & \textbf{61.1} & \textbf{1.52$\times$} \\
    \bottomrule
  \end{tabular}
  }
\end{table}

\textbf{Remark.} Extensive experiments on ImageNet and various downstream pixel-level tasks well demonstrate our superiority in terms of accuracy, efficiency and transferability. Guided by triple-win compression principles, our method successfully deliver accelerated models with high accuracy, fast inference speed, and favorable transferability all at once, showing promising performance in practical scenarios. 

\begin{table}[t]
  \caption{Ablation study on DeiT-Tiny and DeiT-Small. }
  \label{tab:ablation}
  \setlength\tabcolsep{3pt}%
  \centering
  \resizebox{\linewidth}{!}{
  \begin{tabular}{l |ccc|ccc}
    \toprule
      Method & \multicolumn{3}{c|}{DeiT-Tiny} & \multicolumn{3}{c}{DeiT-Small}  \\
      \cmidrule{2-7}
      & \makecell{Top-1} & \makecell{Param.} & \makecell{FLOPs $\downarrow$} & \makecell{Top-1} & \makecell{Param.} & \makecell{FLOPs $\downarrow$} \\
      \midrule
      Original & 72.2\% & 5.7M & - & 79.8\% & 22.1M & -  \\
      Original-600e & 73.5\% & 5.7M & - & 81.0\% & 22.1M & - \\
      ATME &  71.9\% & 5.9M & 50.2\% & 79.8\% & 22.9M & 54.4\% \\
      CDCP &  71.9\% & 4.2M & 31.2\% & 79.8\% & 13.9M & 38.1\% \\

      \rowcolor{lightgray}
      \textbf{CAIT} &  \textbf{72.3\%} & \textbf{5.1M} & \textbf{50.5\%} & \textbf{80.2\%} & \textbf{18.4M}  & \textbf{54.4\%} \\
    \bottomrule
  \end{tabular}
  }
\end{table}

\subsection{Model Analyses}
\label{sec:analyses}
\subsubsection{Ablation Study} We conduct experiments with DeiT-Tiny and DeiT-Small, following~\cite{wei2023joint,wang2022vtc}. As shown in \Cref{tab:ablation}, compared with original models, our ATME can obtain comparable accuracy while reducing 50.2\% and 54.4\% FLOPs for DeiT-Tiny and DeiT-Small, respectively. The proposed CDCP can obtain sufficient FLOPS reduction as well. These results can demonstrate the effectiveness of ATME and CDCP. We can also observe that, compared with CDCP, our ATME, as a token pruning method, can obtain superior performance. This result is consistent with observations in prior works~\cite{wang2022vtc,yang2021nvit,liang2022not} that for ViT models, compressing tokens can achieve more outcomes than compressing channels. Therefore, in practice, we follow ~\cite{wang2022vtc} to assign more FLOPs reduction ratio on ATME. Specifically, given a desired ratio of overall FLOPs reduction, we first prioritize the strategy of uniformly dividing layers for token pruning, and then adjust the FLOPs reduction ratio of channel pruning to exactly match the target overall FLOPs reduction. Furthermore, it can be observed that compared with ATME and CDCP, the final model, CAIT, achieves the best performance. This is attributed to that CAIT can simultaneously eliminate the data level redundancy by ATME and model level redundancy by CDCP, achieving optimal outcomes. Besides, we also continue training the pretrained DeiT-Tiny and DeiT-Small for 300 epochs under the same setting, denoted as ``Original-600e'', which leads to the same epochs as compressing pretrained models and serves as the performance upper bounds. Compared with them, we note that our CAIT also shows competitive performance after significant computation reduction. These results demonstrate the effectiveness and superiority of CAIT.

\begin{table}[t]
\centering
\caption{Comparison with alternative methods on DeiT-Small (Top-1: 79.8\%).}
\setlength\tabcolsep{4pt}%
\label{tab:atme-cdcp}
\resizebox{\linewidth}{!}{
\begin{tabular}[t]{ccccc}
    \toprule
    \makecell{Token Pruning} & \makecell{Channel Pruning} & \makecell{Top-1 (\%)} & \makecell{FLOPs ($\downarrow$\%)} & \makecell{Speed $\uparrow$}  \\
    \midrule
    EViT &- & 79.6 & 43.3\% & 1.7$\times$ \\
    LFE &- & 80.1 & 43.3\% & 1.5$\times$ \\
    \rowcolor{lightgray}
    \textbf{ATME} &-& \textbf{80.0} & \textbf{43.3\%} & \textbf{1.8$\times$} \\
    \cmidrule{1-5}
    -& NViT & 78.9 & 32.8\% & 1.2$\times$ \\
    -& LFS & 79.4 & 32.8\% & 1.2$\times$ \\
    \rowcolor{lightgray}
    -& \textbf{CDCP} & \textbf{79.8} & \textbf{32.8\%} & \textbf{1.2$\times$} \\
    \cmidrule{1-5}
    LFE & LFS & 79.1 & 55.0\% & 1.6$\times$ \\
    LFE & CDCP & 79.6 & 55.0\% & 1.6$\times$ \\
    ATME & LFS & 79.1 & 55.0\% & 2.0$\times$ \\
    \rowcolor{lightgray}
    \textbf{ATME} & \textbf{CDCP} & \textbf{79.5} & \textbf{55.0\%} & \textbf{2.0$\times$} \\
    \bottomrule
\end{tabular}
}
\end{table}
  
\begin{table}[t]
  \centering
  \caption{Impact of asymmetry in ATME with DeiT-Tiny (Top-1: 72.2\%).}
  \label{tab:atme}
  \resizebox{\linewidth}{!}{
  \begin{tabular}[t]{l |cccc}
      \toprule
      Method & \makecell{Top-1 (\%)} & \makecell{Param. (M)} & \makecell{FLOPs ($\downarrow$\%)} & \makecell{Speed $\uparrow$} \\
      \midrule
      symmetry &  71.2 & 6.0 & 50.7\% & 1.9$\times$  \\
      HTM &  71.5 & 5.9 & 49.6\% & 1.8$\times$ \\
      VTM &  71.5 & 5.9 & 49.6\% & 1.8$\times$ \\
      diag & 71.2 & 5.9 & 49.2\% & 1.8$\times$ \\
      \rowcolor{lightgray}
      \textbf{ATME} &  \textbf{71.9} & 5.9 & \textbf{50.2\%} & \textbf{1.9$\times$}  \\
      \bottomrule
  \end{tabular}
  }
\end{table}

\subsubsection{Superiority to Alternative Methods}
To verify the superiority of our proposed ATME and CDCP over existing token pruning and channel pruning methods, we conduct experiments on ImageNet with only compressing tokens, channels, and both. Following \cite{wang2022vtc}, we introduce two state-of-the-art token pruning methods, \ie, EViT~\cite{liang2022not} and LFE~\cite{wang2022vtc}, and two state-of-the-art channel pruning methods, \ie,  NViT~\cite{yang2021nvit} and LFS~\cite{wang2022vtc}, on DeiT-Small, as baselines for ATME and CDCP, respectively. When compressing both tokens and channels, we select better token pruning baseline method, \ie, LFE~\cite{wang2022vtc} and better channel pruning baseline method, \ie, LFS~\cite{wang2022vtc} for combinations. Results of compared baselines are borrowed from \cite{wang2022vtc} directly. For fair comparison, we employ the same training setting as \cite{wang2022vtc}. 

As shown in \Cref{tab:atme-cdcp}, our ATME outperforms EViT by 0.4\% Top-1 accuracy under the same FLOPs reduction. Compared with LFE, our ATME achieves significantly faster inference speed while obtaining comparable accuracy. For channel pruning, our CDCP can outperform NViT and LFS by 0.9\% and 0.4\% accuracy gains, respectively.
When compressing both tokens and channels, our joint compression method is still superior to other combinations, \ie, ATME+LFS, LFE+CDCP, and LFE+LFS. Overall, our ATME and CDCP show their effectiveness compared with other token pruning and channel pruning methods, respectively.

\subsubsection{Asymmetry in ATME}
We investigate the beneficial impacts of asymmetry in ATME. We introduce four baselines:
\begin{enumerate*}[label=(\arabic*)]
  \item simultaneously using horizontal and vertical token merging as one operation, denoted as ``symmetry'', in which we group and concatenate four adjacent tokens in both horizontal and vertical directions, \ie, in a 2$\times$2 patch;
  \item only using horizontal token merging;
  \item only using vertical token merging.
  \item diagonally merging tokens, denoted as ``diag''.
\end{enumerate*}
As shown in \Cref{tab:atme}, ATME can obtain better performance. Specifically, compared with ``symmetry'', ATME progressively reduces the number of tokens in a moderate way, forbidding drastic losses of token information, thus achieving a 0.7\% accuracy gain. Compared with HTM and VTM, ATME can maintain a more regular spatial structure for patches, resulting in a 0.4\% performance improvement. Compared with ``diag'', ATME can enjoy more locality inductive bias and obtain the improvement of 0.7\% accuracy. These results show favorable advantage of asymmetry in ATME.

\begin{table}[t]
\centering
\caption{Impact of consistencies in CDCP with DeiT-Tiny (Top-1: 72.2\%).}
\label{tab:cdcp}
\resizebox{\linewidth}{!}{
\begin{tabular}[t]{l |cccc}
    \toprule
    Method & \makecell{Top-1 (\%)} & \makecell{Param. (M)} & \makecell{FLOPs ($\downarrow$\%)} & \makecell{Speed $\uparrow$} \\
    \midrule
    S$^2$ViTE~\cite{chen2021chasing} &  70.1 & 4.2 & 23.7\%  & 1.1$\times$  \\
    w/o both &  71.0 & 4.3 & 25.1\% & 1.2$\times$  \\
    w/o head &  71.3 & 4.4 & 25.1\%  & 1.2$\times$  \\
    w/o attn &  72.1 & 4.4 & 25.1\%  & 1.2$\times$ \\
    \rowcolor{lightgray}
    \textbf{CDCP} & \textbf{72.7} & 4.5 & \textbf{25.1\%}  & \textbf{1.2$\times$} \\
    \bottomrule
\end{tabular}
}
\end{table}

\subsubsection{Consistencies in CDCP} We verify the positive effects of head-level consistency and attention-level consistency used in CDCP. Additionally, we introduce S$^2$ViTE~\cite{chen2021chasing} as the baseline method, because it is a remarkable state-of-the-art dynamic channel pruning method. As shown in \Cref{tab:cdcp}, head-level and attention-level consistencies can consistently achieve performance improvements. Specifically, head-level consistency leads to a 1.4\% (CDCP 72.7\% vs ``w/o head'' 71.3\%) accuracy gain. Attention-level consistency obtains a 0.6\% (CDCP 72.7\% vs ``w/o attn'' 72.1\%) performance improvement. Besides, our CDCP significantly outperforms the baseline ``w/o both'' and S$^2$ViTE~\cite{chen2021chasing}. Such improvements can be attributed to the fine-grained compression with head-level and attention-level consistencies for ViTs.

\begin{table}[t]
  \centering
  \caption{Results on LV-ViT and Swin Transformer.}
  \label{tab:lvvit}
  \resizebox{\linewidth}{!}{
  \begin{tabular}{cccc}
      \toprule
      Method & \makecell{Top-1 (\%)} & \makecell{FLOPs ($\downarrow$\%)} & \makecell{ Speed $\uparrow$ } \\
      \midrule
      LV-ViT-S~\cite{jiang2021all} & 83.2 & -  & 1.0$\times$ \\ 
      NViT~\cite{yang2021nvit}+EViT~\cite{liang2022not} & 81.5 & 49.2\% & 1.8$\times$ \\
      VTC-LFC~\cite{wang2022vtc} & 81.8 & 50.8\% & 1.2$\times$ \\ 
      \textbf{CAIT} & \textbf{82.2} & \textbf{53.8\%} & \textbf{1.9$\times$} \\ 
      \midrule
      Swin-Tiny~\cite{liu2021swin} & 81.1 & - & 1.0$\times$\\
      SPViT~\cite{he2021pruning} & 80.1 & 24.4\% & - \\
      VTC-LFC~\cite{wang2022vtc} & 80.3 & 26.7\% & 1.2$\times$ \\
      \rowcolor{lightgray}
      \textbf{CAIT} & \textbf{80.6} & \textbf{26.7\%} & \textbf{1.2$\times$} \\
      \bottomrule
  \end{tabular}
  }
  \end{table}

\subsubsection{Compression on Other ViT Models} 
\label{sec:swin}
To explore the performance of our method on other variants of ViTs, we conduct experiments on LV-ViT~\cite{jiang2021all} and Swin Transformer~\cite{liu2021swin}.
Following \cite{wang2022vtc}, we adopt ATME and CDCP on LV-ViT, and employ CDCP to Swin.
Meanwhile, for simplicity, the proposed token labels in the original LV-ViT are not used during training, like~\cite{wang2022vtc}.
As shown in \Cref{tab:lvvit}, our method can consistently achieve the state-of-the-art performance on both models. Specifically, on LV-ViT, our method outperforms VTC-LFC~\cite{wang2022vtc} with 0.4\% higher accuracy while achieving significantly faster acceleration (CAIT 1.9$\times$ vs VTC-LFC 1.2$\times$). For Swin Transformer, our compressed model can also obtain accuracy gains of 0.5\% and 0.3\% compared with SPViT~\cite{he2021pruning}/VTC-LFC~\cite{wang2022vtc}, respectively, under the similar FLOPs reduction ratio. These results well demonstrate the generalization of our method on other ViT variants. Besides, we can observe that LV-ViT and Swin Transformer generally suffer more accuracy drop after pruning than DeiT, which is consistent with previous works~\cite{wang2022vtc}. We hypothesize the reason lies in the architectural differences among LV-ViT, Swin Transformer, and DeiT. Specifically, LV-ViT adopts a narrower expansion ratio in FFN and a deeper layout to improve efficiency. It also leverages token labeling to introduce individual location-specific supervision. Swin Transformer adopts the hierarchical structure and shifted window design to enhance efficiency. Therefore, LV-ViT and Swin Transformer exhibit less data-level (\ie, tokens) redundancy and model-level (\ie, parameters) redundancy, compared with DeiT. They thus suffer more accuracy drop after pruning. Additionally, our proposed method can consistently outperform existing methods on LV-ViT and Swin Transformer, showing promising performance for pruning various ViTs.

\subsection{Discussion}
\label{sec:discussion}

\subsubsection{Locality Matters for ATME}
We provide more insightful analyses for ATME. The proposed ATME uniformly aggregates features of neighboring tokens, which can be regarded as a general architecture for modern ViTs. Thus, we construct a vision transformer model whose architecture is the same as our ATME. Then, we train this model for 600 epochs from scratch with hard distillation of the pretrained model. We denote this model as ``ATME-scratch''. Besides, we introduce two additional models whose FLOPs are similar to ATME-scratch's. One involves halving the depth of DeiT-Tiny, which reduces the number of blocks. The other involves halving the width of DeiT-Tiny, which reduces the embedding dimension. We denote these two models as ``DeiT-half depth'' and ``DeiT-half dim'', respectively, which are trained under the same setting as ``ATME-scratch''. We compare these three models with the one obtained by our ATME pruning method. As shown in \Cref{tab:architecture}, ``ATME-scratch'' significantly outperforms ``DeiT-half depth'' and ``DeiT-half dim'' by 7.1\% and 4.8\% in terms of Top-1 accuracy, respectively. This may be attributed to the inductive bias of locality introduced by our ATME strategy. We also incorporate the extra convolution into the HTM and VTM by appending the depthwise convolution with kernel size of 3$\times$3, which is denoted as ``ATME+Conv''. Compared with ATME, it brings negligible performance gain due to the inherent locality in ATME. Moreover, ``ATME-scratch'' is inferior to DeiT-Tiny by a great margin of 1.1\% accuracy. In contrast, our ATME can result in only a 0.3\% drop, compared with vanilla DeiT-Tiny, while achieving a superior speedup of 1.9$\times$. It indicates that in addition to introducing the locality, our ATME can further preserve the pretrained model's ability to capture visual features and prevent knowledge forgetting during pruning. Thanks to them, our ATME can well serve as a compression methodology for ViTs, delivering high performance and fast inference speed.

\begin{table}[t]
  \small
  \centering
  \caption{Impact of different usages of ATME.}
  \label{tab:architecture}
  \setlength\tabcolsep{4pt}%
  \resizebox{\linewidth}{!}{
  \begin{tabular}[t]{l|cccc}
    \toprule
    Method & \makecell{Top-1 (\%)} & \makecell{Param. (M)} & \makecell{FLOPs (G)} & \makecell{ Speed $\uparrow$ } \\
    \midrule
    DeiT-Tiny & 72.2 &5.7 & 1.3  & 1.0$\times$ \\
    DeiT-half depth & 64.0 & 3.0 & 0.6 & 1.9$\times$ \\
    DeiT-half dim & 66.3 & 2.8 & 0.6 & 1.2$\times$ \\
    ATME-scratch & 71.1 &5.9 & 0.6  & 1.9$\times$ \\
    ATME+Conv & 71.9 &6.0 & 0.6  & 1.9$\times$ \\
    \rowcolor{lightgray}
    \textbf{ATME} & \textbf{71.9} &5.9 & 0.6  & \textbf{1.9$\times$} \\
    \bottomrule
  \end{tabular}
  }
\end{table}

\begin{table}[t]
  \small
  \centering
  \caption{ATME in Hierarchical Architectures.}
  \label{tab:swin}
  \resizebox{\linewidth}{!}{
  \begin{tabular}[t]{l|cccc}
    \toprule
    Method & \makecell{Top-1 (\%)} & \makecell{Param. (M)} & \makecell{FLOPs ($\downarrow$\%)} & \makecell{ Speed $\uparrow$ } \\
    \midrule
    Swin-Tiny~\cite{liu2021swin} & 81.1 & 28.3 & - & 1.0$\times$ \\
    CDCP & 80.6 & 24.6 & 26.7 & 1.2$\times$ \\
    \rowcolor{lightgray}
    \textbf{CAIT} & \textbf{80.5} & 25.5 & \textbf{35.4} & \textbf{1.3}$\times$ \\
    \bottomrule
  \end{tabular}
  }
\end{table}

\begin{table}[t]
  \small
  \centering
  \caption{Comparison with ToMe and DiffRate on ImageNet. * indicates that ToMe is adopted in the same pruning layers as our ATME.}
  \label{tab:tome}
  \resizebox{\linewidth}{!}{
  \begin{tabular}[t]{l|cccc}
    \toprule
    Method & \makecell{Top-1 (\%)} & \makecell{Param. (M)}  & \makecell{FLOPs ($\downarrow$\%)} & \makecell{ Speed $\uparrow$ } \\
    \midrule
    DeiT-Small & 79.8 & 22.1 & - & 1.0$\times$ \\
    ToMe & 79.9 & 22.1 & 41.3  & 1.5$\times$ \\
    ToMe+Param & 80.0 & 22.6 & 42.7  & 1.5$\times$ \\
    ToMe* & 80.0 & 22.1 & 41.6 & 1.7$\times$ \\
    DiffRate & 80.1 & 22.1 & 41.3 & 1.6$\times$ \\
    \rowcolor{lightgray}
    \text{ATME} & \textbf{80.0} & 22.6 & \textbf{43.3} & \textbf{1.8$\times$} \\
    \bottomrule
  \end{tabular}
  }
\end{table}

\begin{table}[t]
  \small
  \centering
  \caption{Comparison with ToMe and DiffRate on ADE20k using Semantic FPN. * indicates that ToMe is adopted in the same pruning layers as our ATME.}
  \label{tab:tome_ade}
  \resizebox{\linewidth}{!}{
  \begin{tabular}[t]{l|cccc}
    \toprule
    Method & \makecell{mIoU} & \makecell{Param. (M)}  & \makecell{FLOPs ($\downarrow$\%)} & \makecell{ Speed $\uparrow$ } \\
    \midrule
    DeiT-Small & 44.3 & 22.1 & - & 1.0$\times$ \\
    ToMe & 44.0 & 22.1 & 41.3  & 1.5$\times$ \\
    ToMe+Param & 44.0 & 22.6 & 42.7  & 1.5$\times$ \\
    ToMe* & 43.9 & 22.1 & 41.6 & 1.7$\times$ \\
    DiffRate & 43.1 & 22.1 & 41.6 & 1.6$\times$ \\
    \rowcolor{lightgray}
    \textbf{ATME} & \textbf{44.5} & 22.6 & \textbf{43.3} & \textbf{1.8}$\times$ \\
    \bottomrule
  \end{tabular}
  }
\end{table}

\subsubsection{ATME in Hierarchical Architectures} As an efficient token pruning strategy for DeiT, our proposed ATME can also transfer to hierarchical architectures, \eg, Swin Transformer. We conduct experiments under the same setting in \Cref{sec:swin} to verify this. Specifically, we adopt HTM and VTM at the last two layers of the 3-th Stage in Swin-Tiny, respectively, which results in a FLOPs reduction of 20.1\%. We further perform channel pruning on it, leading to an overall 35.4\% FLOPs reduction. As shown in \Cref{tab:swin}, CAIT obtains a comparable accuracy with only performing channel pruning on Swin-Tiny, but with a much larger FLOPs reduction (35.4\% vs. 26.7\%) and a more significant inference speedup (1.3x vs. 1.2x). It well demonstrates the effectiveness of our token compression method in transferring to hierarchical architectures.

\subsubsection{Superiority of ATME to Others}
ToMe~\cite{bolya2022token} and DiffRate~\cite{chen2023diffrate} are existing state-of-the-art token pruning methods, which leverage bipartite soft matching to merge similar tokens. To demonstrate the superiority of our proposed ATME for token pruning, we compare our strategy with three variants:  
\begin{enumerate*}[label=(\arabic*)] 
  \item using the strategy in ToMe in the same pruning layers as our ATME; and
  \item performing ToMe at every layer as in their paper~\cite{bolya2022token}; and
  \item adopting the DiffRate for token pruning.
\end{enumerate*} Besides, we also introduce extra parameters to ToMe like ours. To maintain the similar number of parameters and the FLOPs reduction, we incorporate the \textsf{Linear} layer after the merging of ToMe in every three blocks and slightly increase the number of merged tokens, which is denoted as ``ToMe+Param''.

We first conduct experiments on ImageNet under the same experimental setups to investigate their performance based on DeiT-Small. Specifically, we finetune ToMe and DiffRate under the same training setting as ours. As shown in \Cref{tab:tome}, our ATME obtains a comparable accuracy with the ToMe variants and DiffRate under a larger FLOPs reduction, demonstrating the effectiveness of our asymmetric token merging method. Besides, the strategy in ToMe and DiffRate employs the complex bipartite similarity matching with complex operators, while our ATME simply merges neighboring tokens and utilizes fast tensor manipulations. Our ATME thus affords a significant advantage for various devices and platforms, particularly those with limited computation ability or lacking support for complex operators. As evidenced in \Cref{tab:tome}, our ATME is more friendly to latency and leads to an advantageous inference speedup compared with ToMe and DiffRate.

More importantly, the strategy in ToMe and DiffRate merges tokens sparsely at each pruning layer, resulting in the disruption of the spatial integrity of images and restricting the transferability of compressed models to downstream structured vision tasks. In contrast, our ATME can well preserve the complete structure of patches and maintain the strong transferability of ViTs. We further conduct experiments on the downstream semantic segmentation task to verify this, by Semantic FPN segmentation method. To transfer the strategy in ToMe and DiffRate to the downstream task, we track which tokens get merged and then unmerge tokens, \ie, using the merged token to fill the corresponding empty positions, when constructing feature maps to be fed into the segmentation decoder. Besides, for DiffRate which also sparsely drops tokens, we adopt the same strategy as ``VTC-LFC-structured'' to obtain the complete feature map. As shown in \Cref{tab:tome_ade}, since DiffRate's pruning strategy is specifically tuned on ImageNet, its performance significantly degrades when transferred to the downstream task. Besides, our ATME outperforms others with considerable margins, along with a larger inference speedup, showing the favorable transferability for downstream structured vision tasks.

\begin{table}[t]
  \small
  \centering
  \caption{Comparison with STViT-R on ImageNet. $\dagger$ means the reproduced performance using the official code.}
  \label{tab:stvit}
  \resizebox{\linewidth}{!}{
  \begin{tabular}[t]{l|cccc}
    \toprule
    Method & \makecell{Top-1 (\%)} & Param.(M) & \makecell{FLOPs ($\downarrow$\%)} & \makecell{ Speed $\uparrow$ } \\
    \midrule
    Swin-Tiny & 81.3 & 28.3 & - & 1.0$\times$ \\
    STViT-R & 80.5$^\dagger$ & 28.3 & 19.6 & 1.2$\times$ \\
    \rowcolor{lightgray}
    \text{ATME} & \textbf{81.1}& 28.9 & \textbf{20.1} & \textbf{1.2$\times$} \\
    \midrule
    Swin-Small & 83.2 & 49.6 & - & 1.0$\times$ \\
    STViT-R & 82.5$^\dagger$/82.7 & 49.6 & 33.0 & 1.3$\times$ \\
    \rowcolor{lightgray}
    \text{ATME} & \textbf{83.0}& 50.2 & \textbf{33.7} & \textbf{1.3$\times$} \\
    \bottomrule
  \end{tabular}
  }
\end{table}

\begin{table}[t]
  \small
  \centering
  \caption{Comparison with STViT-R on object detection (AP$^{box}$) and instance segmentation (AP$^{mask}$) on COCO, and semantic segmentation (mIoU) on ADE20k.}
  \label{tab:stvit_coco}
  \resizebox{\linewidth}{!}{
  \begin{tabular}[t]{l|cc|c|c}
    \toprule
    Method & AP$^{box}$ & \makecell{AP$^{mask}$} & mIoU & \makecell{ Speed $\uparrow$ } \\
    \midrule
    Swin-Tiny & 50.5 & 43.7 & 45.8 & 1.0$\times$ \\
    STViT-R & 49.4$^\dagger$ & 42.7$^\dagger$ & 43.9$^\dagger$ & 1.2$\times$ \\
    \rowcolor{lightgray}
    \textbf{ATME} & \textbf{50.1} & \textbf{43.3} & \textbf{45.2}  & \textbf{1.2$\times$} \\
    \midrule
    Swin-Small & 51.8 & 44.7 & 49.5 & 1.0$\times$ \\
    STViT-R & 51.6$^\dagger$/51.8 & 44.7$^\dagger$/44.7 & 46.4$^\dagger$/48.3 & 1.3$\times$ \\
    \rowcolor{lightgray}
    \textbf{ATME} & \textbf{51.8} & \textbf{44.7} & \textbf{48.5}  & \textbf{1.3$\times$} \\
    \bottomrule
  \end{tabular}
  }
\end{table}

\begin{table}[t]
  \small
  \centering
  \caption{Comparison with STViT-R on medical image segmentation.}
  \label{tab:stvit_medical}
  \begin{tabular}[t]{l|ccccc}
    \toprule
    Method & \makecell{Synapse} & \makecell{ACDC} & \makecell{Polyp} & \makecell{ Speed $\uparrow$ } \\
    \midrule
    Swin-Tiny & 79.5 & 90.0 & 80.6  & 1.0$\times$ \\
    STViT-R & 79.0 & 89.7 & 79.8 & 1.2$\times$ \\
    \rowcolor{lightgray}
    \textbf{ATME} & \textbf{79.3} & \textbf{90.1} & \textbf{80.5} & \textbf{1.2$\times$} \\
    \bottomrule
  \end{tabular}
\end{table}

\begin{table}[t]
  \small
  \centering
  \caption{Comparison with STViT-R on aerial semantic segmentation.}
  \label{tab:stvit_aerial}
  \setlength\tabcolsep{7pt}%
  \begin{tabular}[t]{l|cc|c|cc}
    \toprule
    \multirow{2}{*}{Method} & \multicolumn{2}{c|}{Potsdam} & iSAID & \multirow{2}{*}{Speed $\uparrow$} \\
    \cmidrule{2-4}
    & OA & mF1 & mIoU \\
    \midrule
    Swin-Tiny & 91.2 & 90.6 & 64.6  & 1.0$\times$ \\
    STViT-R & 90.9 & 90.4 & 63.2 & 1.2$\times$  \\
    \rowcolor{lightgray}
    \textbf{ATME} & \textbf{91.1} & \textbf{90.6} & \textbf{64.6} & \textbf{1.2$\times$} \\
    \bottomrule
  \end{tabular}
\end{table}

We further compare our ATME with STViT-R~\cite{chang2023making}, which presents the recovery module and dumbbell unit to perform token pruning and adapt to downstream tasks. Note that its another variant STViT hinders the application for downstream tasks~\cite{chang2023making} due to the side
effect of losing nearly all the detailed information, we thus leave out it. We follow STViT-R to leverage ATME on Swin Transformer and train from scratch for 300 epochs. To achieve comparable FLOPs reduction with STViT-R, we adopt the HTM and VTM at the 4-th and 5-th layers of the third stage for Swin-Tiny, and at the 6-th and 12-th layers of the third Stage for Swin-Small, respectively. As shown in \Cref{tab:stvit}, on ImageNet, our ATME significantly outperforms STViT-R by 0.6\% and 0.5\% top-1 accuracies on Swin-Tiny and Swin-Small, respectively. It shows the favorable high accuracy of ATME. Besides, we follow STViT-R to transfer the models to object detection and instance segmentation using Cascade Mask R-CNN, and semantic segmentation using UperNet. All the training settings follow STViT-R. As shown in \Cref{tab:stvit_coco}, our ATME surpasses STViT-R by 0.7 AP$^{box}$ and 0.6 AP$^{mask}$ on Swin-Tiny. Additionally, ATME outperforms STViT-R by considerable margins on semantic segmentation. Due to that STViT-R only remains the tokens with high-level semantic information, it loses nearly all the detailed pixel-level information and thus suffers inferior performance on downstream dense vision tasks. In contrast, our ATME can well maintain the complete spatial structure and dense position information, showing strong transferability. Furthermore, we also compare ATME and STViT-R on out-of-domain downstream tasks, \ie, medical image segmentation and aerial semantic segmentation. We adopt the same experimental setups as \Cref{sec:medical}. As shown in \Cref{tab:stvit_medical}, our ATME outperforms STViT-R by 0.7 DICE score on Polyp dataset. As shown in \Cref{tab:stvit_aerial}, ATME obtains 1.4 mIoU improvement over STViT-R on iSAID. These results demonstrate robust generalizability of ATME over STViT-R on downstream vision tasks.

\subsubsection{Robustness of CDCP to Compression Schedule} We follow \cite{ding2021resrep} to set the compression training schedule of CDCP. To verify that the pruning performance of our CDCP is not sensitive to different compression schedules, we conduct experiments on DeiT-Tiny to analyze the effects of the epoch for starting increasing $r_{target}$ and interval iterations for the reconstruction of $P$. As shown in \Cref{tab:warmup} and \Cref{tab:interval}, we can see that they do not make significant differences, indicating the robustness of CDCP. Therefore, the effectiveness of our method is general and not limited by specific schedules.

\begin{table}[t]
  \small
  \centering
  \caption{Results for different epochs for starting increasing $r_{target}$ on DeiT-Tiny.}
  \label{tab:warmup}
  \setlength\tabcolsep{7pt}%
  \resizebox{\linewidth}{!}{
  \begin{tabular}[t]{c|cccc}
    \toprule
    warmup epochs & \makecell{Top-1 (\%)}  & \makecell{FLOPs ($\downarrow$\%)} & Speed $\uparrow$  \\
    \midrule
    0 & 72.26 & 50.5 & 1.7$\times$  \\
    15 & 72.31 & 50.5 & 1.7$\times$  \\
    30 & 72.34 & 50.5 & 1.7$\times$  \\
    \bottomrule
  \end{tabular}
  }
\end{table}

\begin{table}[t]
  \small
  \centering
  \caption{Results for different interval iterations for the reconstruction of $P$ on DeiT-Tiny.}
  \label{tab:interval}
  \setlength\tabcolsep{7pt}%
  \resizebox{\linewidth}{!}{
  \begin{tabular}[t]{c|cccc}
    \toprule
    interval iterations & \makecell{Top-1 (\%)}  & \makecell{FLOPs ($\downarrow$\%)} & Speed $\uparrow$ \\
    \midrule
    25 & 72.34 & 50.5 & 1.7$\times$ \\
    50 & 72.37 & 50.5 & 1.7$\times$ \\
    75 & 72.46 & 50.5 & 1.7$\times$ \\
    \bottomrule
  \end{tabular}
  }
\end{table}

\begin{table}[!t]
  \small
  \centering
  \caption{Results for different fine-tuning epochs on DeiT-Small. The suffix "-Xe" means X epochs.}
  \label{tab:epochs}
  \setlength\tabcolsep{8pt}%
  \resizebox{\linewidth}{!}{
  \begin{tabular}[t]{c|cccc}
    \toprule
    Method & Top-1 (\%) & FLOPs ($\downarrow$\%) & Speed $\uparrow$ \\
    \midrule
    DeiT-Small & 79.8 & - & 1.0$\times$  \\
    CAIT-30e & 76.7 & 55.0 & 2.0$\times$ \\
    CAIT-150e & 79.5 & 55.0 & 2.0$\times$ \\
    CAIT-300e & 80.2 & 55.0 & 2.0$\times$ \\
    \bottomrule
  \end{tabular}
  }
\end{table}

\subsubsection{Different Fine-tuning Epochs for CAIT}
To investigate the performance of different fine-tuning epochs of our CAIT, we conduct experiments on DeiT-Small. As shown in \Cref{tab:epochs}, due to introducing parameters for extra optimization, our method benefits more from longer fine-tuning epochs, and results in a high performance upper bound. Specifically, our CAIT-150e and CAIT-300e enjoys a 2.0$\times$ inference speedup with no or little accuracy drops.

\begin{figure*}
  \centering
  \includegraphics[width=0.8\linewidth]{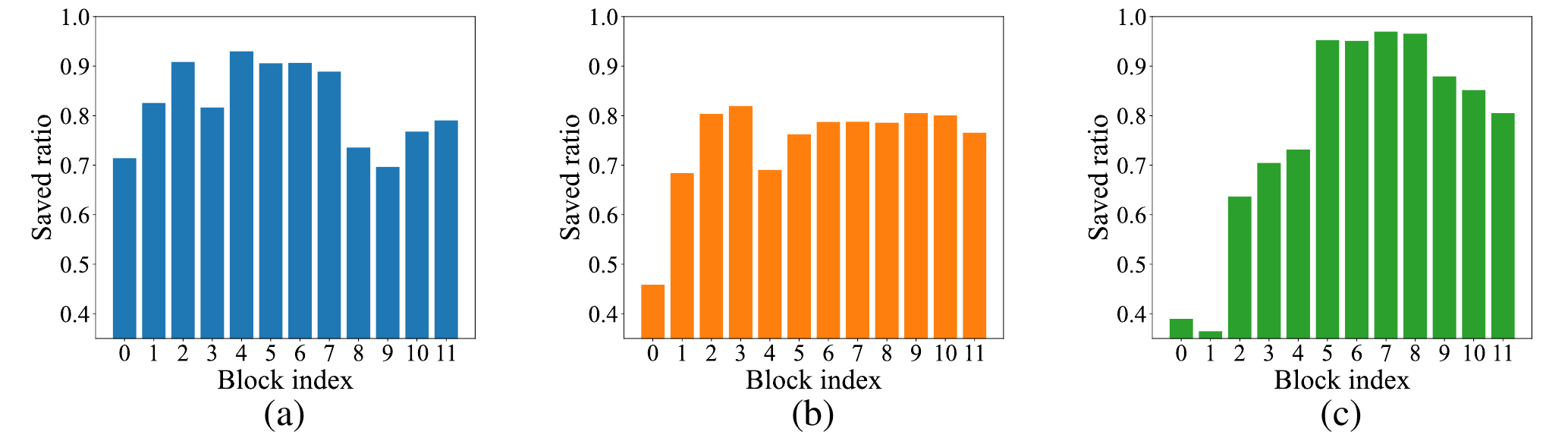}
  \caption{Architecture of pruned (a) DeiT-Tiny, (b) DeiT-Small, and (c) DeiT-Base models.}
  \label{fig:pruned}
  \vspace{-10pt}
\end{figure*}

\begin{figure*}
  \centering
  \includegraphics[width=0.9\linewidth]{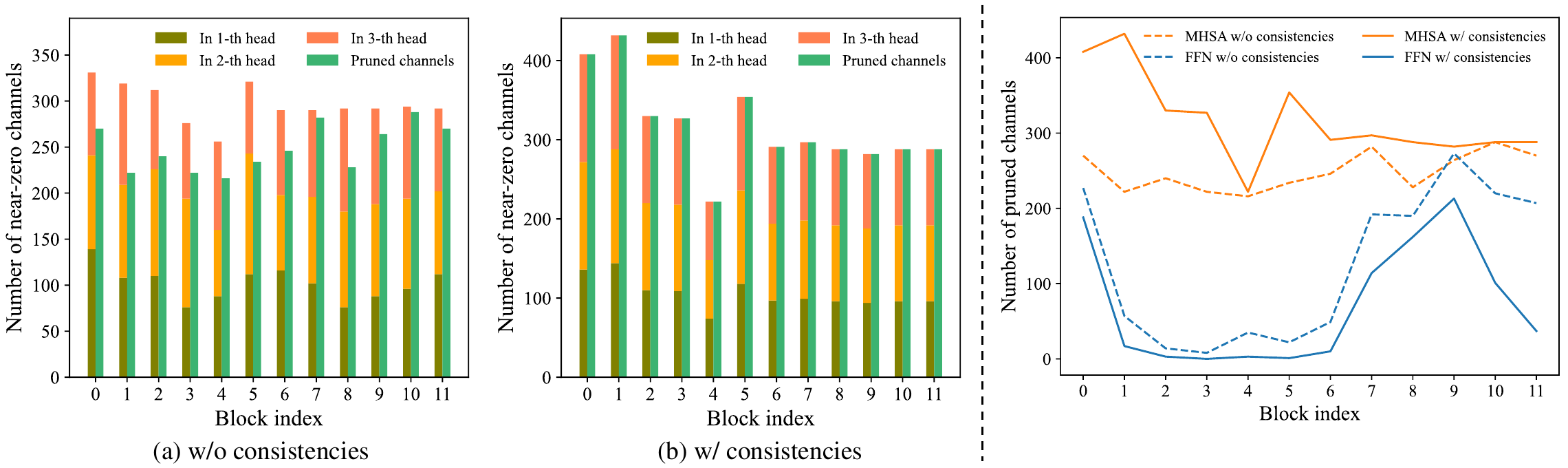}
  \caption{Near-zero importance channels VS. Pruned channels in MHSA (left). Pruned channels in MHSA and FFN (right).}
  \label{fig:consistency}
  \vspace{-15pt}
\end{figure*}

\subsubsection{Parameter Distribution of Pruned Models}
We visualize the parameter distribution of pruned DeiT-Tiny, DeiT-Small and DeiT-Base. \Cref{fig:pruned} presents the saved ratios of channels in each block. It reveals that middle and deep blocks tend to retain more channels than shallow blocks, which is consistent with observations in prior works~\cite{wang2022vtc,zheng2022savit}. This phenomenon may be attributed to the fact that middle and deep blocks incorporate more global context and thus capture more complex visual representations. Furthermore, it may provide some insight towards the construction of efficient ViTs. For example, we can maintain narrower channels in shallow blocks of ViTs.

\subsubsection{Visualization of Consistencies}
\label{sec:cdcp_vis}
We conduct visualization analyses to show the positive effects of our proposed head-level consistency and attention-level consistency in CDCP. Specifically, we visualize the near-zero importance channels in each head and the ultimately pruned channels in the MHSA module, based on the DeiT-Tiny model with three heads. As mentioned in \Cref{sec:cdcp}, directly applying the conventional compactor pruning strategy for ViTs will cause imbalance ratios of pruned channels among heads and inconsistent pruned channels between query and key transformation matrices, leading to difficulties for parallel and error-free self-attention computation. Then, only the minimum ratio of near-zero importance channels among heads can be pruned and only the consistent near-zero importance channels between query and key transformation matrices can be removed. However, as shown in \Cref{fig:consistency}.(a), such a strategy will cause a substantial number of channels close to zero importance are retained in the compressed model, which impacts the performance adversely (\Cref{tab:cdcp}). In contrast, our head-level and attention-level consistencies can maintain different heads of the same block in the same shape, and well align remained channels in the query and key, respectively. Therefore, as shown in \Cref{fig:consistency}.(b), our proposed consistencies can well address the limitations of vanilla compactor pruning on ViTs, ensuring error-free parallel self-attention computation and leading to superior performance (\Cref{tab:cdcp}). 
Besides, as shown in \Cref{fig:consistency}, we can also observe that our proposed consistencies result in more pruned channels in MHSA modules and less pruned channels in FFN modules. This may be attributed to the fact that we encourage consistent shapes of different heads and aligned channels of the query and key in MHSA during pruning. It is also consistent with observations in previous works~\cite{yang2021nvit,liu2023efficientvit,yu2022metaformer} that more redundant channels lie in MHSA modules. Additionally, results in Table 5 in the paper demonstrate the effectiveness of such a pruning strategy.

\begin{figure*}
  \centering
  \includegraphics[width=0.9\linewidth]{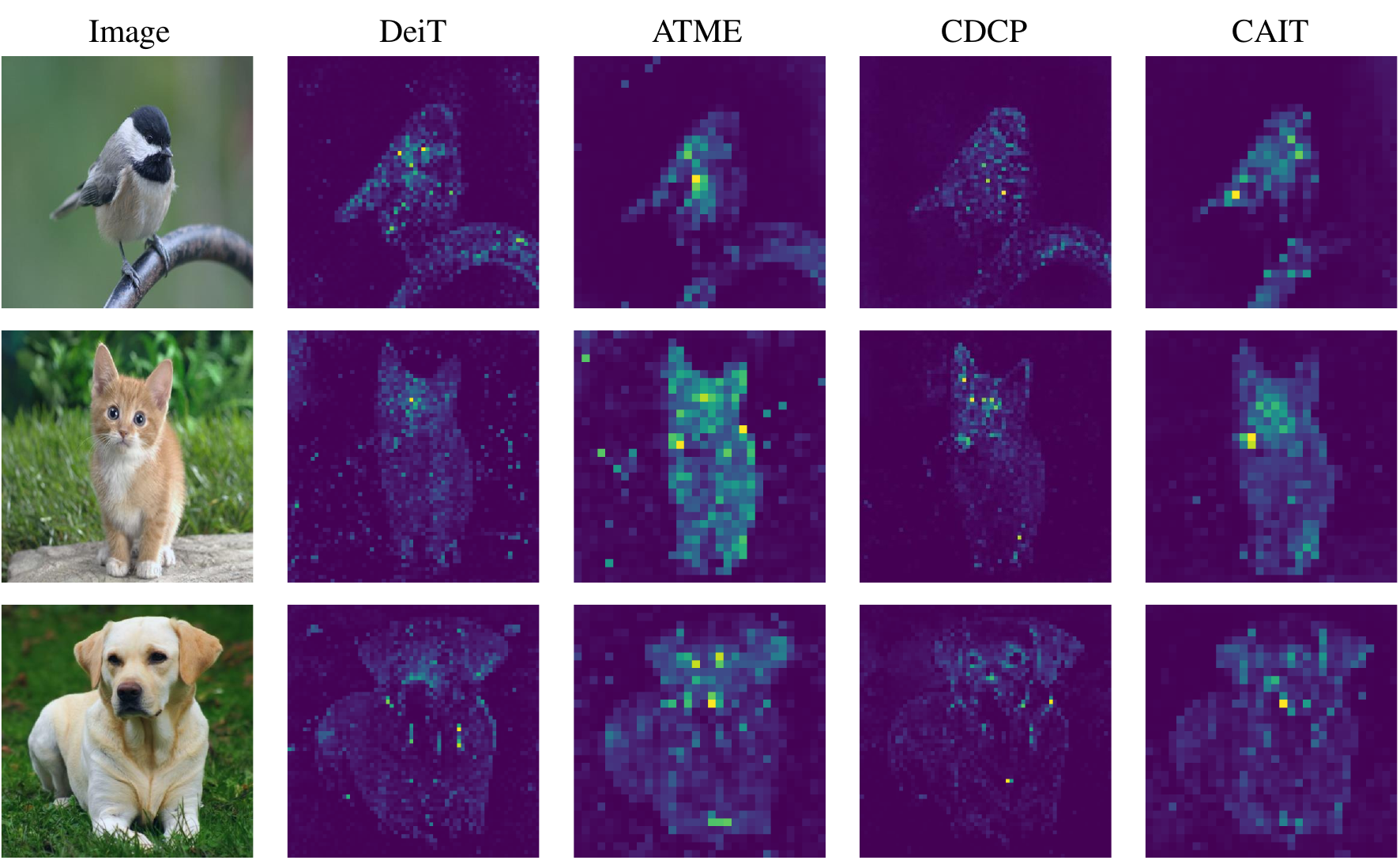}
  \caption{Attention map comparison with token pruning by ATME, channel pruning by CDCP, and the joint pruning by CAIT.}
  \label{fig:attn}
\end{figure*}

\subsubsection{Visualization of Attention Maps}
We conduct visualization analyses to inspect the impact caused by token pruning and channel pruning for attention maps based on DeiT-Small. Following~\cite{caron2021emerging}, we visualize the attention map of the \texttt{[CLS]} token at the last layer. As shown in \Cref{fig:attn}, the attention maps can be well preserved after token pruning by ATME, channel pruning by CDCP, and the joint pruning by CAIT. Besides, after token pruning, the important visual areas with highly semantic information can be strengthened due to the eliminated data level (\ie, tokens) redundancy. Furthermore, thanks to the reduced model level (\ie, parameters) redundancy by CDCP, the noise in the less informative regions can be well suppressed due to the less disturbance during computing attention. These favorable properties well help model to grasp critical visual information better, leading to improved efficiency.

\section{Conclusion}
\label{sec:conclusion}
In this paper, we proposes CAIT, a joint compression method with asymmetric token merging and consistent dynamic channel pruning for ViTs. The proposed asymmetric token merging strategy can effectively reduce the number of tokens while maintaining the spatial structure of images. The consistent dynamic channel pruning strategy can perform dynamic fine-grained compression optimization for all modules in ViTs. Extensive experiments on multiple ViTs over various vision tasks show that our method can outperform state-of-the-arts, achieving high performance, fast inference speed, and favorable transferability at the same time, well demonstrating its effectiveness and superiority.

\textbf{Limitations.} Although our CAIT shows superior performance and efficiency, it falls short of original models under certain scenarios and fails to achieve lossless compression. Besides, its transferability to 3D tasks and video tasks is also worth exploring. We leave these for future work.

\textbf{Acknowledgement.} This work was supported by National Natural Science Foundation of China (Nos. 62525103, 624B2082, 62271281, 62441235, 62571294).

\bibliography{IEEEabrv,tpami.bib}
\bibliographystyle{IEEEtran}

\section{Biography Section}

\vspace{-33pt}

\begin{IEEEbiography}[{\includegraphics[width=1in,height=1.25in,clip,keepaspectratio]{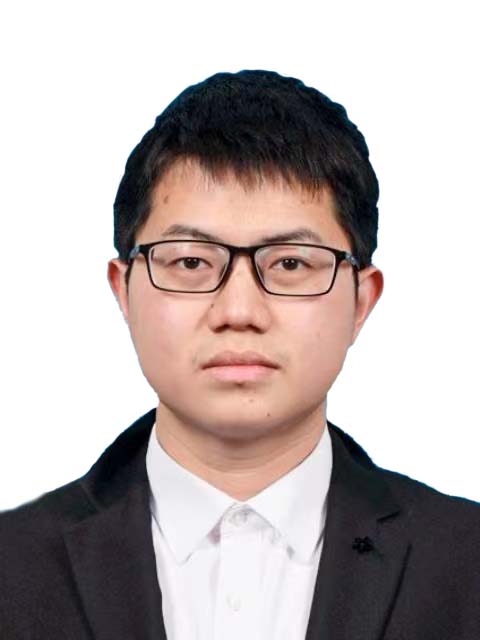}}]{Ao Wang}received the BE degree in the School of Software, Tsinghua University, China, in 2022. He is currently working toward the PhD degree in the School of Software with Tsinghua University, China. His research interests include computer vision and machine learning.
\end{IEEEbiography}
\vspace{-35pt}
\begin{IEEEbiography}[{\includegraphics[width=1in,height=1.25in,clip,keepaspectratio]{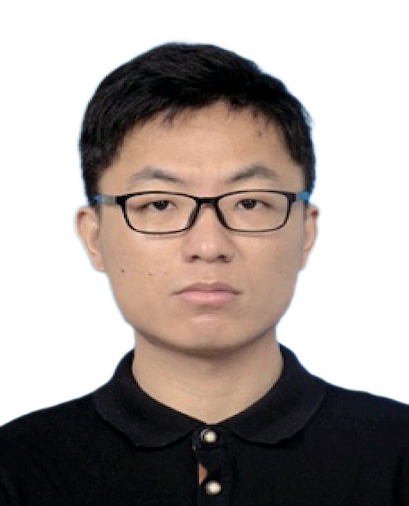}}]{Hui Chen} is currently an assistant researcher with Beijing National Research Center for Information Science and Technology, Tsinghua University. His research interests include efficient and effective multi-modal perception and learning. He has published more than 15 peer-reviewed top
conference and journal papers, including CVPR, ICCV, ICLR, etc. He served as a PC member of several top-tier conferences.
\end{IEEEbiography}
\vspace{-35pt}
\begin{IEEEbiography}[{\includegraphics[width=1in,height=1.25in,clip,keepaspectratio]{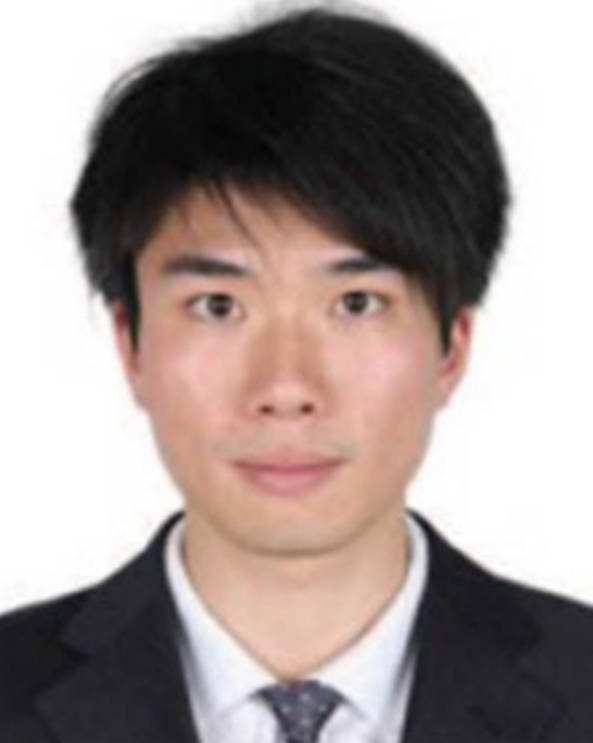}}]{Zijia Lin} received the B.Sc. degree from the School of Software, Tsinghua
University, Beijing, China, in 2011, and the Ph.D. degree from the Department of Computer Science and Technology, Tsinghua University, in 2016. His research interests include multimedia information retrieval and machine learning.
\end{IEEEbiography}
\vspace{-35pt}
\begin{IEEEbiography}[{\includegraphics[width=1in,height=1.25in,clip,keepaspectratio]{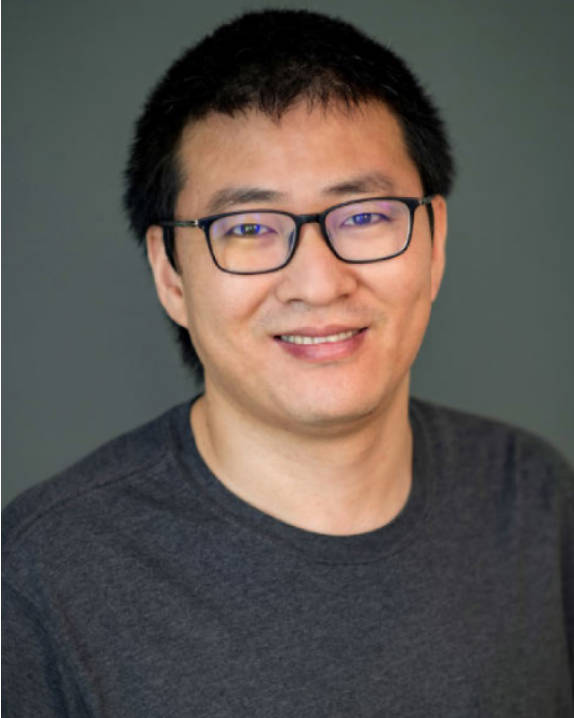}}]{Sicheng Zhao} received the Ph.D. degree from the Harbin Institute of Technology, Harbin, China, in 2016. He was a Visiting Scholar with the National University of Singapore, Singapore, from 2013 to 2014; a Research Fellow with Tsinghua University, Beijing, China, from 2016 to 2017; a Postdoctoral Research Fellow with the University of California at Berkeley, Berkeley, CA, USA, from 2017 to 2020; and a Postdoctoral Research Scientist with Columbia University, New York, NY, USA, from 2020 to 2022. He is
currently a Research Associate Professor with Tsinghua University. His research interests include affective computing, multimedia, and computer vision.
\end{IEEEbiography}
\vspace{-35pt}
\begin{IEEEbiography}[{\includegraphics[width=1in,height=1.25in,clip,keepaspectratio]{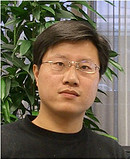}}]{Jungong Han} is currently a tenured professor in the Department of Automation at Tsinghua University. He also holds an Honorary Professorship at the University of Warwick, UK. He has authored 2 edited volumes, and over 200 papers, including 90 in prestigious IEEE/ACM Transactions, and 60+ in CORE A* conferences. He is a Fellow of IAPR and a Fellow of AAIA. His research interests include computer vision and multi-modal learning.
\end{IEEEbiography}
\vspace{-35pt}
\begin{IEEEbiography}[{\includegraphics[width=1in,height=1.25in,clip,keepaspectratio]{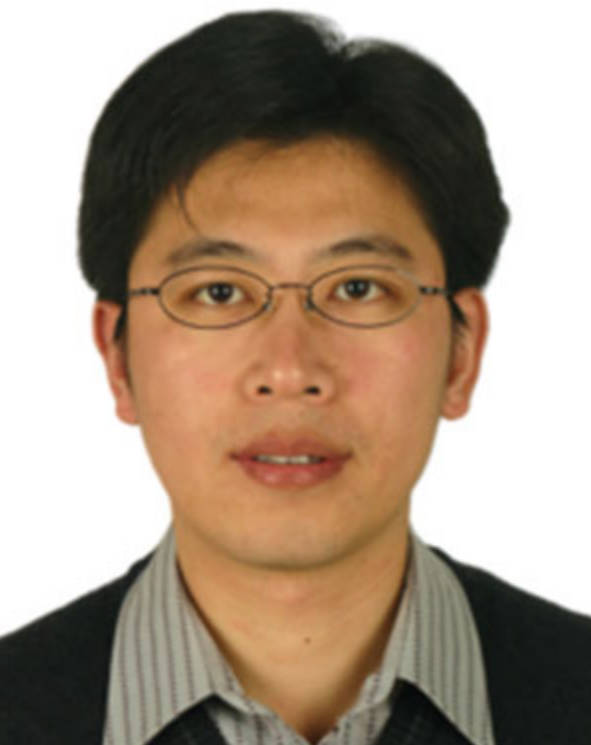}}]{Guiguang Ding} is currently a tenured professor in the School of Software at Tsinghua University. He has dedicated himself to promising fields, such as model architecture design and compression, visual semantic recognition and description, transfer learning, and few-shot learning. He has over 30 papers published in top-tier journals including IEEE TPAMI, IEEE SPM, and IEEE TIP. Additionally, he has presented more than 70 papers at top-tier international conferences, such as CVPR, NeurIPS, ICML, etc.
\end{IEEEbiography}

\vfill

\end{document}